\documentclass{article}

\usepackage[final,eandd]{neurips_2026}

\usepackage[utf8]{inputenc}
\usepackage[T1]{fontenc}
\usepackage{url}
\usepackage{booktabs}
\usepackage{multirow}
\usepackage{graphicx}
\usepackage{subcaption}
\usepackage{xcolor}
\usepackage{hyperref}
\hypersetup{hidelinks}
\usepackage{amsmath}
\usepackage{amsfonts}
\usepackage{nicefrac}
\usepackage{microtype}
\usepackage[most]{tcolorbox}
\usepackage{pifont}  % For \ding symbols (checkmarks, crosses)
\usepackage{placeins}  % For \FloatBarrier
\definecolor{darkgreen}{rgb}{0.0, 0.55, 0.0}
\usepackage{tikz}
\usetikzlibrary{positioning}

\usepackage{enumitem}
\setlist[itemize]{leftmargin=*}
\newcommand{\flybench}{\textsc{FlyAOC}}

\definecolor{taskblue}{HTML}{3498DB}
\definecolor{taskgreen}{HTML}{27AE60}
\definecolor{taskpurple}{HTML}{9B59B6}
\newcommand{\taskgo}{\textcolor{taskblue}{\textsc{gene ontology}}}
\newcommand{\taskexpr}{\textcolor{taskgreen}{\textsc{expression}}}
\newcommand{\tasksyn}{\textcolor{taskpurple}{\textsc{synonyms}}}
\newcommand{\tgo}{\textsc{gene ontology}}
\newcommand{\texpr}{\textsc{expression}}
\newcommand{\tsyn}{\textsc{synonyms}}

\newcommand{\tablefont}{\fontsize{8pt}{9pt}\selectfont}
\newcommand{\densetablefont}{\fontsize{7pt}{8pt}\selectfont}
\newcommand{\compacttablespacing}{\setlength{\tabcolsep}{4.5pt}\renewcommand{\arraystretch}{0.95}}
\newcommand{\densetablespacing}{\setlength{\tabcolsep}{3.5pt}\renewcommand{\arraystretch}{0.92}}
\newcommand{\tablenote}[1]{\par\vspace{0.25em}\begin{minipage}{0.95\linewidth}\fontsize{7pt}{8pt}\selectfont #1\end{minipage}}

\begin{document}

%%
%% Title
%%
\title{\flybench{}: Evaluating Agentic Ontology Curation of Drosophila Scientific Knowledge Bases}

%%
%% Authors
%%
\author{%
\parbox[t]{0.29\textwidth}{\centering\normalfont
\textbf{Xingjian Zhang} \\
\small University of Michigan \\
\texttt{jimmyzxj@umich.edu}
}
\And
\parbox[t]{0.29\textwidth}{\centering\normalfont
\textbf{Sophia Moylan} \\
\small University of Illinois\\
Urbana-Champaign \\
\texttt{smoylan2@illinois.edu}
}
\And
\parbox[t]{0.29\textwidth}{\centering\normalfont
\textbf{Ziyang Xiong} \\
\small University of California,\\
Berkeley \\
\texttt{xziyang@berkeley.edu}
}
\AND
\parbox[t]{0.29\textwidth}{\centering\normalfont
\textbf{Qiaozhu Mei} \\
\small University of Michigan \\
\texttt{qmei@umich.edu}
}
\And
\parbox[t]{0.29\textwidth}{\centering\normalfont
\textbf{Yichen Luo} \\
\small University of Washington \\
\texttt{yichenl@uw.edu}
}
\And
\parbox[t]{0.29\textwidth}{\centering\normalfont
\textbf{Jiaqi W. Ma} \\
\small University of Illinois\\
Urbana-Champaign \\
\texttt{jiaqima@illinois.edu}
}
}

\maketitle

%%
%% Abstract
%%
\begin{abstract}
%%
%% Abstract
%%
Scientific knowledge bases accelerate discovery by curating findings from primary literature into structured, queryable formats for both human researchers and emerging AI systems. Maintaining these resources requires expert curators to search papers, reconcile evidence across documents, and produce ontology-grounded annotations. Existing benchmarks usually evaluate isolated subtasks, such as named entity recognition or relation extraction, and therefore do not capture this end-to-end workflow. We present \flybench{} to evaluate AI agents on end-to-end \underline{a}gentic \underline{o}ntology \underline{c}uration from scientific literature. Given a gene symbol, a concise FlyBase gene description, access to a 16,898-paper corpus, and ontology resources, agents must search for evidence and recover as many curator-relevant structured annotations as possible. Outputs span standardized function terms, expression patterns, and historical synonyms linking decades of nomenclature. The benchmark includes 7,397 expert-curated annotations across 100 genes drawn from FlyBase, the \textit{Drosophila} knowledge base. Across four baseline agent harnesses---memorization, fixed pipeline, single-agent, and multi-agent---\flybench{} is sensitive to harness design, model family, and tool-use reliability. These results reveal system-level failure modes that model-only evaluations do not capture.
% Our analysis surfaces several findings to guide future development; for example, agents primarily use retrieval to confirm parametric knowledge rather than discover new information. 
\flybench{} provides a reproducible testbed for retrieval-augmented scientific curation.\footnote{Code: \url{https://github.com/xingjian-zhang/flyaoc}; Data: \url{https://huggingface.co/datasets/anonymous-042/flyaoc}.}

\end{abstract}

%%
%% Overview figure
%%
\begin{figure}[t]
\centering
\begin{subfigure}[t]{0.32\textwidth}
    \centering
    \includegraphics[width=\linewidth,height=4.6cm,keepaspectratio]{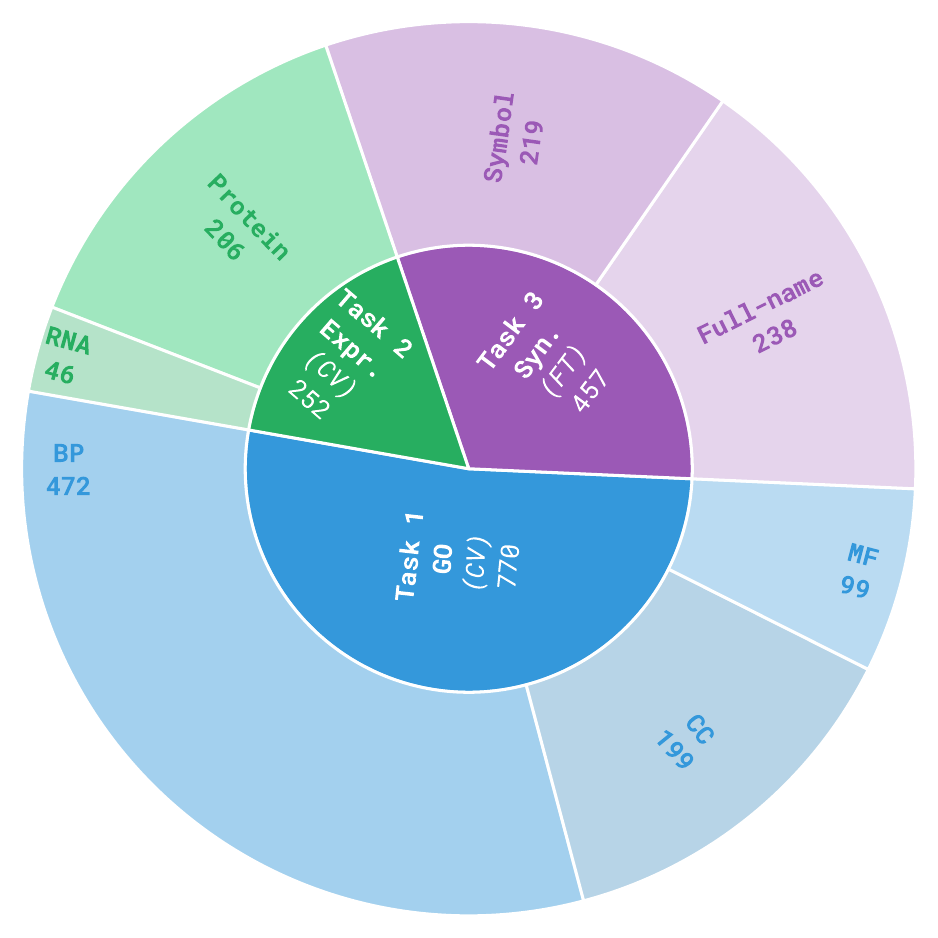}
    \caption{Task counts.}
    \label{fig:task-sunburst}
\end{subfigure}
\hfill
\begin{subfigure}[t]{0.64\textwidth}
    \centering
    \includegraphics[width=\linewidth,height=4.6cm,keepaspectratio]{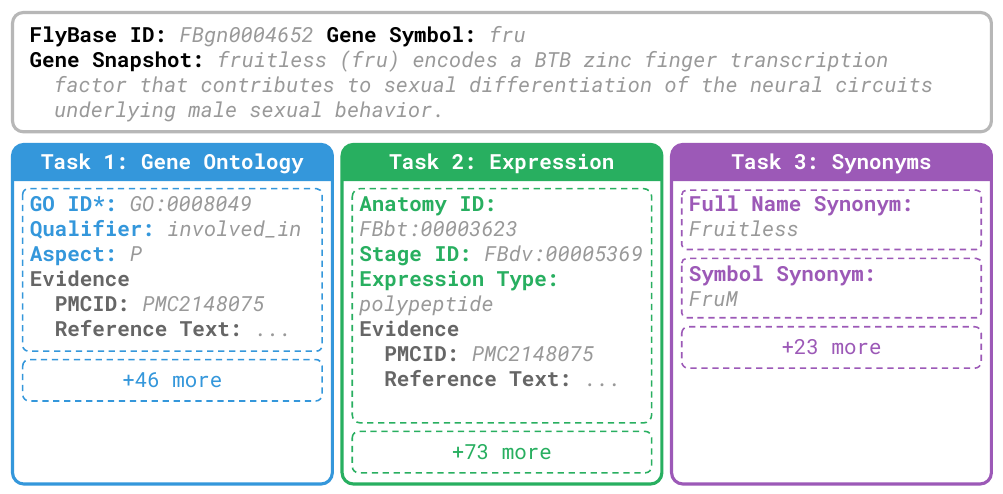}
    \caption{Example ground truth for gene \textit{fruitless} (fru).}
    \label{fig:data-example}
\end{subfigure}
\caption{\flybench{} overview. Given a gene symbol, a concise gene description from FlyBase, and access to 16,898 full-text papers and ontology files, agents produce structured annotations for \textit{Drosophila melanogaster} genes. (a) Three tasks: \taskgo{}, \taskexpr{}, and \tasksyn{}; counts show deduplicated verified facts used for primary evaluation. Tasks 1--2 use controlled vocabularies (CV); Task 3 uses free text (FT). (b) Example ground truth for \textit{fruitless}; badges indicate additional annotations beyond those shown.}
\label{fig:tasks-overview}
\end{figure}

%% Main sections
%%
%% 1. Introduction
%%
\section{Introduction}
\label{sec:intro}

Scientific knowledge bases are a foundational resource for modern research. Databases in biology~\citep{uniprot2025uniprot, ashburner2000gene, thurmond2019flybase, lamesch2012arabidopsis}, chemistry~\citep{Groom_2016}, and materials science~\citep{Jain_2013} organize decades of findings into structured, queryable formats. This organization accelerates discovery across disciplines. These resources also increasingly support AI systems. Retrieval-augmented approaches that ground language model outputs in curated knowledge reduce hallucinations, improve factual accuracy, and provide access to domain knowledge without reprocessing primary literature~\citep{li2024simple, manda2025llms}.

These resources are mainly produced through hybrid manual--automatic curation pipelines. Expert curators systematically review primary literature to extract structured annotations, including standardized labels for gene functions (``Gene Ontology'', or GO). Automated tools assist with subtasks such as named-entity recognition, relation extraction, and ontology term normalization~\citep{gene2017expansion}. However, the rapidly expanding scientific literature exceeds the practical limits of manual curation. Careful paper-by-paper analysis has historically restricted curators at major model organism databases, including TAIR and FlyBase~\citep{lamesch2012arabidopsis, thurmond2019flybase}, to only a small portion of the available evidence~\citep{li2012building, McQuilton_2012}. As a result, databases increasingly rely on automated pipelines to supplement manual curation~\citep{du2011and, uniprot2025uniprot}.

Large language models (LLMs) offer a promising path toward automating aspects of curation and literature review~\citep{scherbakov2025emergence, cai2025utilizing}. However, assessing whether these capabilities transfer to real-world curation tasks remains challenging because few benchmarks evaluate LLMs on ontology-grounded curation tasks. Existing curation benchmarks typically focus on isolated skills---named entity recognition, relation extraction, or question answering over pre-selected passages~\citep{Van_Auken_2014}. These evaluations are valuable, but they do not capture the end-to-end workflow that curators actually perform: given a topic of interest, find relevant papers from a large corpus, read and comprehend full-text articles, and produce structured annotations grounded in domain ontologies.

We introduce \textbf{\flybench{}}, a benchmark for \underline{a}gentic \underline{o}ntology \underline{c}uration based on \underline{Fly}Base~\citep{thurmond2019flybase}. Unlike prior work that provides curated text spans for extraction~\citep{Van_Auken_2014}, \flybench{} gives agents a gene symbol, a concise gene description from FlyBase (the FlyBase Gene Snapshot), and access to 16,898 full-text scientific papers plus ontology files. Agents must navigate this literature and produce structured annotations, matching the setting in which curation tools would be deployed. \flybench{} directly targets \textbf{knowledge base curation}. It also evaluates agents' capability for \textbf{in-depth literature search and synthesis} in a specialized scientific domain, capabilities central to many AI-for-science applications.

\flybench{} focuses on \textit{Drosophila melanogaster} (fruit fly), which combines a century of genetic research with a gold-standard curated knowledge base (FlyBase~\citep{thurmond2019flybase}). The benchmark methodology can transfer to any organism database, or any scientific domain, with curated annotations and accessible literature. Drosophila provides a strong test bed because it has both high-quality annotations and a large literature. We define three complementary curation tasks---extracting gene functions, expression patterns, and historical synonyms---that together characterize a gene's biological identity and exercise different retrieval and synthesis capabilities.

We evaluate four baseline agent harnesses representing different points in the design space: a \emph{Memorization} baseline using only parametric knowledge stored in LLMs, a fixed \emph{Pipeline} with parallel document processing, and two autonomous agents (\emph{Single-Agent} and \emph{Multi-Agent}) with iterative tool use. Here, a harness is the execution scaffolding around a language model, including prompts, tools, orchestration, context management, validation, and feedback loops. Our experiments show that literature-based curation depends on the coupled model--harness configuration. Context management changes how performance scales with retrieved papers, and cross-family comparisons reveal substantial differences in model capability, tool-use reliability, and cost.

\flybench{} offers several properties that distinguish it from prior benchmarks:
\begin{itemize}
    \item \textbf{Agent-oriented}: The benchmark evaluates end-to-end curation and literature search, from retrieval through structured output, rather than isolated NLP components.
    \item \textbf{Large-scale}: With 16,898 full-text papers totaling over 140 million words, \flybench{} is large enough to stress retrieval, prioritization, and context management.
    \item \textbf{Corpus-grounded}: Not all expert annotations are derivable from our corpus. We identify recoverable labels through evidence-recoverability verification over curator-linked sources, so evaluation focuses on annotations that agents could plausibly extract from the provided literature.
    \item \textbf{Extensible}: We release the complete corpus, evaluation framework, and baseline implementations to support development of improved methods.
\end{itemize}

Together, these design choices make \flybench{} a reproducible benchmark for agentic ontology curation from scientific literature.

%%
%% 2. Related Work
%%
\section{Related Work}
\label{sec:related}

\subsection{Agent Benchmarks and Scientific Literature}

Recent benchmarks have established rigorous evaluation for AI agents across domains: SWE-bench tests autonomous software engineering over real GitHub issues~\citep{jimenez2024swebench}, while WebArena evaluates web navigation through realistic browser tasks~\citep{zhou2024webarena}. These benchmarks share a design philosophy: agents receive high-level goals, execute multi-step workflows, and are evaluated against ground-truth outcomes rather than intermediate steps. \flybench{} applies this paradigm to scientific literature. Agents receive a gene symbol and a concise gene description, then use corpus and ontology resources to produce structured annotations evaluated against expert-curated ground truth.

Scientific literature presents distinct challenges. Unlike web tasks with observable state or code with executable tests, extracting knowledge from papers requires comprehending dense technical prose, synthesizing information across documents, and grounding outputs in domain ontologies. Deep research agents such as OpenAI Deep Research, Gemini Deep Research, and Perplexity have begun addressing these challenges in user-facing products~\citep{openai2025deepresearch, google2024deepresearch, perplexity2025deepresearch}, retrieving and synthesizing information from hundreds of online sources into consolidated reports. PaSa~\citep{he2025pasa} targets academic paper search specifically, training an agent to iteratively refine queries and rank results, though it evaluates retrieval without downstream extraction.

Several benchmarks evaluate such systems on question answering (HLE~\citep{phan2025humanity}, GAIA~\citep{mialon2024gaia}) or report generation (DeepResearch Bench~\citep{du2025deepresearchbench}, ReportBench~\citep{li2025reportbench}). These settings emphasize different targets. Question answering tests whether a system can locate a target answer and stop. Report generation tests broader synthesis, but is difficult to score objectively for completeness and faithfulness. \textbf{\flybench{} evaluates a complementary capability: high-recall structured candidate generation over verified corpus-grounded annotations.} Given a gene, agents recover curator-relevant annotations across dozens of papers and return a ranked structured record for bounded review. This emphasizes recall-oriented coverage rather than single-answer accuracy or free-text synthesis, while retaining tractable evaluation through ontology-grounded outputs. Recent agentic frameworks that do adopt ontology-grounded extraction, such as STRUCTSENSE~\citep{chhetri2025structsense}, evaluate performance on single documents or individual artifacts. \flybench{}, by contrast, evaluates corpus-scale synthesis.

\subsection{Biomedical Information Extraction}

A long line of work has developed systems for extracting structured biological knowledge from literature. Early approaches used pattern matching~\citep{muller2004textpresso} and supervised learning for entity disambiguation~\citep{he2010bsqa}. Subsequent work expanded supervised learning for text extraction~\citep{gobeill2012answering}, explored merging document- and topic-level prediction~\citep{lena2015gota}, and developed sophisticated rule-based relation extraction systems~\citep{ravikumar2017belminer}. More recently, LLM-based systems such as SPIRES apply zero-shot extraction to organize text according to ontology schemas~\citep{caufield2023structured}, and tools like CurateGPT combine LLMs with retrieval and ontology search to assist human curators~\citep{caufield2024curategpt}. Domain-specific applications have demonstrated LLM-based extraction of synthesis parameters~\citep{zheng2023chatgpt} and protein-ligand bioactivity data~\citep{yan2025biominer} from individual or pre-selected papers.

Despite these advances, rigorous evaluation remains limited. Existing biomedical benchmarks emphasize isolated component-level tasks---entity recognition, term normalization, or relation extraction---evaluated independently rather than as part of integrated workflows. BC4GO, for instance, frames Gene Ontology curation as identifying GO terms within 200 pre-selected articles~\citep{Van_Auken_2014}; CRAFT provides rich annotations over 97 full-text documents~\citep{bada2012concept}. GOTA evaluates Gene Ontology (GO) term assignment from biomedical literature using a benchmark of 15,000 annotated publications~\citep{lena2015gota}. These benchmarks enable valuable component evaluation, but they do not test whether systems can autonomously navigate from a high-level query to structured output.

A common thread across prior work is the separation of retrieval from extraction: search systems evaluate query ranking without downstream extraction~\citep{he2025pasa}, and extraction benchmarks evaluate annotation on pre-selected documents~\citep{Van_Auken_2014, zheng2023chatgpt, yan2025biominer}. \textbf{\flybench{} evaluates the \emph{coupled} retrieval-extraction process end-to-end.} This coupling introduces challenges absent from component evaluation: agents receive no relevance guarantee for retrieved papers, retrieval errors propagate directly into extraction failures, and effective curation requires adapting search strategy based on what has already been found. The 16,898-paper corpus exposes prioritization and context-management problems that smaller corpora cannot probe.

%%
%% 3. Benchmark Tasks
%%
\section{Benchmark Tasks}
\label{sec:tasks}

\subsection{Task Inputs and Available Resources}
\label{sec:tasks:input}

We adopt a gene-centric formulation: given a gene, synthesize relevant annotations from the literature. This reflects how researchers query biological databases and how FlyBase gene pages are organized~\citep{McQuilton_2012}. It complements the paper-centric workflow of curators, who triage new publications and extract annotations for all genes mentioned.

For each gene, agents receive a gene symbol, a concise gene description from FlyBase (the Gene Snapshot), access to full-text papers, and ontology files (Figure~\ref{fig:data-example}). This expert-written description provides biological context without exposing target labels for function, expression, or synonyms. The corpus contains papers associated with benchmark genes (Section~\ref{sec:dataset}); ontology files let agents search and validate controlled-vocabulary terms.

\subsection{Task Outputs}
\label{sec:tasks:output}

Agents produce three types of structured annotations for each gene (Figure~\ref{fig:task-sunburst}). Task~1 asks \emph{what does the gene do?}, requiring Gene Ontology terms. Task~2 asks \emph{where and when is it expressed?}, requiring anatomy and developmental stage terms. Its primary score evaluates the anatomy component because anatomy is usually the more informative expression target and has a more granular ontology; developmental-stage IDs remain part of the output schema but are not included in the canonical reported metrics. Task~3 asks \emph{what other names has it been called?}, requiring synonyms. Together, these tasks represent common queries that geneticists have about genes.

The tasks require retrieval, extraction, and aggregation rather than single-answer lookup: a gene can have many relevant functions, expression contexts, and historical names spread across multiple papers. Synonyms add complexity because historical names can unlock additional literature. Agents must \textbf{rank predictions by confidence}, enabling curators to prioritize verification.

\noindent\textbf{Task 1: \taskgo{}.}
Task~1 requires Gene Ontology (GO) term identifiers (e.g., \texttt{GO:0007399}) describing biological roles. GO organizes terms into Biological Process, Molecular Function, and Cellular Component~\citep{ashburner2000gene,GeneOntology_2023}. Its directed acyclic graph enables partial-credit semantic evaluation (Appendix Figure~\ref{fig:go-hierarchy}; Section~\ref{sec:tasks:eval}).

\noindent\textit{Missing-term setting of Task 1.}
We also define a missing-term setting for Task~1 to evaluate concepts that lack ontology support. This simulates curation workflows where findings appear before formal ontology terms. We hide rare GO leaf terms that appear in only one corpus paper, serving as a proxy for novel concepts.

\noindent\textbf{Task 2: \taskexpr{}.}
Task~2 requires (anatomy term, developmental stage) tuples describing where and when the gene is expressed. For primary evaluation, each verified anatomy term is treated as the item to recover. Developmental stages remain part of the output schema for evidence context and future diagnostic evaluation. Annotations use the FlyBase Anatomy Ontology and Developmental Ontology~\citep{FlyBase_2024}; the anatomy hierarchy enables semantic scoring.

\noindent\textbf{Task 3: \tasksyn{}.}
Task~3 requires alternative gene names. Over a century of \textit{Drosophila} research, genes were often independently discovered and named by different laboratories; \textit{Notch}, for example, has synonyms including \textit{N}, \textit{notch-1}, \textit{Ax}, and \textit{fa}.

\noindent\textbf{Note.} Our baselines produce all three tasks in one run to share input tokens such as paper content, but the benchmark also supports methods that tackle tasks separately.

\subsection{Evaluation Metrics}
\label{sec:tasks:eval}

We evaluate agents using recall@$k$: agents rank predictions by confidence, and we measure what fraction of ground-truth annotations appear in the top $k$. This matches curation as bounded, high-recall candidate generation: missed annotations force open-ended literature search, while surfaced candidates can be checked by experts. We report primary results on the \textbf{corpus-grounded subset}, whose evidence is recoverable from accessible structured text in the provided corpus (Section~\ref{sec:dataset}). This isolates retrieval and extraction ability from corpus coverage; secondary metric definitions and precision/F1 diagnostics are in Appendix~\ref{app:metrics}.

\noindent\textbf{\taskgo{} and \taskexpr{}: Semantic recall.}
We relax recall@$k$ for ontology tasks by awarding partial credit for related predictions using Wang semantic similarity~\citep{Wang_2007}. For a gene $i$ with ground-truth annotations $\text{GT}_i$ and top-$k$ predictions $\text{Pred}_{i,1:k}$, semantic recall@$k$ is:
\[
\text{Recall}_i@k = \frac{\sum_{g \in \text{GT}_i} \max_{p \in \text{Pred}_{i,1:k}} \text{sim}(g, p)}{|\text{GT}_i|}
\]
where $\text{sim}(g, p) \in [0, 1]$ equals 1 for exact matches and decreases with ontological distance. Task~1 uses the GO DAG. Task~2 primary recall uses the FlyBase Anatomy hierarchy over verified expression anatomy-term annotations.

\noindent\textbf{\tasksyn{}: Exact recall.}
We use case-insensitive exact recall@$k$ for string matching.

\noindent\textbf{Aggregation.}
Primary results use micro-averaged corpus-grounded recall over deduplicated gene-specific evaluation items: GO term annotations, expression anatomy-term annotations, and synonym strings. Failed or empty-output runs are scored as zero recall for that gene rather than excluded. Appendix~\ref{app:metrics} gives aggregation formulas, macro averages, and secondary metrics.

\noindent\textbf{Choice of $k$.}
We set $k$ to roughly four times the mean number of verified facts per gene for each task (Table~\ref{tab:k-choice}). This gives agents room to recover most relevant annotations while keeping candidate review bounded.

\vspace{-10pt}
\begin{table}[ht]
\centering
\tablefont
\compacttablespacing
\caption{Verified Annotations and Choice of $k$.}
\label{tab:k-choice}
\begin{tabular}{lccc}
\toprule
Task & Verified Annotations & Mean/gene & $k$ \\
\midrule
\taskgo{} & 770 & 7.7 & 30 \\
\taskexpr{} & 252 & 2.5 & 10 \\
\tasksyn{} & 457 & 4.6 & 20 \\
\bottomrule
\end{tabular}
\end{table}
\vspace{-10pt}

%%
%% 4. Dataset Construction
%%
\section{Dataset Construction}
\label{sec:dataset}

This section describes how we select benchmark genes, assemble the corpus, and establish ground-truth annotations. Figure~\ref{fig:data-construction} in Appendix~\ref{app:stats} summarizes the pipeline.

\subsection{Gene Selection}
\label{sec:dataset:genes}

We select 100 benchmark genes from FlyBase through a three-stage filter for rich annotations and extractable ground truth. FlyBase catalogs approximately 17,000 protein-coding genes~\citep{FlyBase_2024}, but annotation depth varies considerably. Our filter identifies genes with sustained research interest and sufficient corpus-grounded labels.

\noindent\textbf{Stage 1: Research interest.} We select genes with FlyBase \emph{Gene Snapshots}, which are concise, expert-written gene descriptions indicating sustained research attention, yielding 3,999 candidates.

\noindent\textbf{Stage 2: Literature availability.} We retain genes with at least 10 full-text papers in our corpus, reducing to 3,446 genes.

\noindent\textbf{Stage 3: Corpus-grounded annotations.} We require at least one corpus-grounded annotation for \emph{both} Task~1 (\tgo{}) and Task~2 (\texpr{}), ensuring that every benchmark gene has ground truth an agent could potentially extract. This yields 314 candidate genes. We rank them by the minimum corpus-grounded count across tasks, favoring balanced coverage over genes dominated by a single task, and select the top 100.

\subsection{Literature Corpus}
\label{sec:dataset:corpus}

Our corpus contains 16,898 PMC open-access full-text papers retrieved via the BioC-PMC API from over 105,000 FlyBase-referenced publications~\citep{comeau2019pmc} (Appendix Figure~\ref{fig:corpus-overview}). It provides structured article text, from introduction through discussion, totaling over 140 million words (Appendix~\ref{app:stats}). For post-2000 FlyBase references, coverage is 24\% (16,665 of 69,404), rising to 30--40\% after 2015. This coverage reflects realistic automated-curation access, where paywalled venues remain difficult to retrieve programmatically.

\subsection{Ground Truth and Corpus Grounding}
\label{sec:dataset:ground-truth}

We construct ground truth in two stages. First, we take expert target labels from FlyBase. FlyBase curators record gene functions, expression patterns, and synonyms using structured vocabularies and literature provenance~\citep{FlyBase_2024}, defining the biological targets for \flybench{}. Second, we determine which labels are observable from the released corpus. This step is necessary because some cited evidence is absent from PMC or appears only in figures, supplements, or other non-text content. An agent using the released structured text therefore could not recover every FlyBase label.

To avoid penalizing agents for evidence outside the released text, we first use reference matching and then verify evidence recoverability. For Task~1 and Task~2, reference matching identifies FlyBase labels whose curator-linked publication maps to a PMC paper in our corpus. We then ask GPT-5 whether that paper's accessible structured text supports the known label for the target gene. This bounded setup keeps the judgment tractable: the prompt supplies one rendered FlyBase annotation, including ontology IDs and evidence metadata when available, plus the paper title, abstract, and section text. The model only decides whether the supplied label is recoverable from that text, returning a support verdict and an evidence span when support is found. It does not search the corpus, discover annotations, or normalize ontology terms. Labels with multiple sources are retained if any cited paper is supported. Appendix~\ref{app:grounding-verification} gives the prompt, strictness criteria, and domain-expert spot checks of verifier rationales.

This filter removes labels unavailable to agents, including those supported only by inaccessible papers, non-text modalities, supplements, or computational inference. Task~3 synonyms are grounded by text matching because their evidence is lexical. Primary evaluation deduplicates the retained subset into verified facts (Table~\ref{tab:k-choice}) so recall measures retrieval and extraction from the provided corpus rather than corpus coverage gaps.

%%
%% 5. Experiments
%%
\section{Experiments}
\label{sec:experiments}

We evaluate four baseline agent harnesses on \flybench{}, spanning pure memorization to autonomous tool use. We report corpus-grounded recall@$k$ as the primary metric (Section~\ref{sec:tasks:eval}) with gene-bootstrap intervals over benchmark genes. Appendix~\ref{app:agent-benchmark-practice} discusses this interval choice.

\subsection{\texorpdfstring{Experimental Setup}{Experimental Setup}}
\label{sec:experiments:agents}

\noindent\textbf{Shared tools.} Retrieval-based harnesses use the same \emph{literature tools} for BM25 corpus search and retrieval~\citep{robertson2009probabilistic}, \emph{ontology tools} for GO/anatomy/stage IDs, and a \emph{validation tool} for schema checking.\footnote{The raw corpus and ontology files are included in our benchmark. Researchers can develop custom tools for new agents.} BM25 is deliberately simple and practical; Appendix~\ref{app:retrieval} ablates more complex retrieval alternatives.

\noindent\textbf{Baseline harnesses.} The four harnesses test the main design choices in retrieval-augmented curation:
\begin{itemize}
    \item \textbf{Memorization.} Uses no corpus retrieval. It shares the Single-Agent design otherwise, including ontology tools, and measures what the model can recover from parametric knowledge alone.
    \item \textbf{Pipeline.} Uses a fixed five-stage workflow: search, retrieve top papers, extract natural-language descriptions per paper, batch-resolve ontology IDs, and compile outputs. It is cheap and parallel, but has no feedback loop for query expansion or ontology correction.
    \item \textbf{Single-Agent.} Uses one ReAct-style~\citep{yao2022react} LLM to choose tools iteratively, reformulate searches, and refine ontology lookups. All retrieved paper text accumulates in one context window.
    \item \textbf{Multi-Agent.} Uses an orchestrator that delegates paper reading to subagents. The orchestrator sees extracted annotations rather than raw paper text, keeping context bounded.
\end{itemize}

Table~\ref{tab:agent-comparison} summarizes the comparison. Pipeline, Single-Agent, and Multi-Agent are \emph{retrieval-based methods}; gains over Memorization measure the value of literature access, while gains over Pipeline indicate the value of adaptive agentic workflows.

\vspace{-10pt}
\begin{table}[ht]
\centering
\tablefont
\compacttablespacing
\caption{Comparison between baseline agent harnesses.}
\label{tab:agent-comparison}
\begin{tabular}{lcccc}
\toprule
& \textbf{Retrieval} & \textbf{Feedback loop} & \textbf{Context$^*$} & \textbf{Relative cost} \\
\midrule
Memorization & No & Yes & Accumulated & Low \\
Pipeline & Yes & No & Partitioned & Lowest \\
Single-Agent & Yes & Yes & Accumulated & High \\
Multi-Agent & Yes & Yes & Partitioned & Medium \\
\bottomrule
\end{tabular}
\tablenote{$^*$Whether retrieved paper text accumulates in a single LLM context window or is partitioned across subagents.}
\vspace{-10pt}
\end{table}

\subsection{\texorpdfstring{Experiment 1: Harness Scaling and Cost}{Experiment 1: Harness Scaling and Cost}}
\label{sec:experiments:harnesses}

\begin{figure*}[!t]
    \centering
    \includegraphics[width=\textwidth]{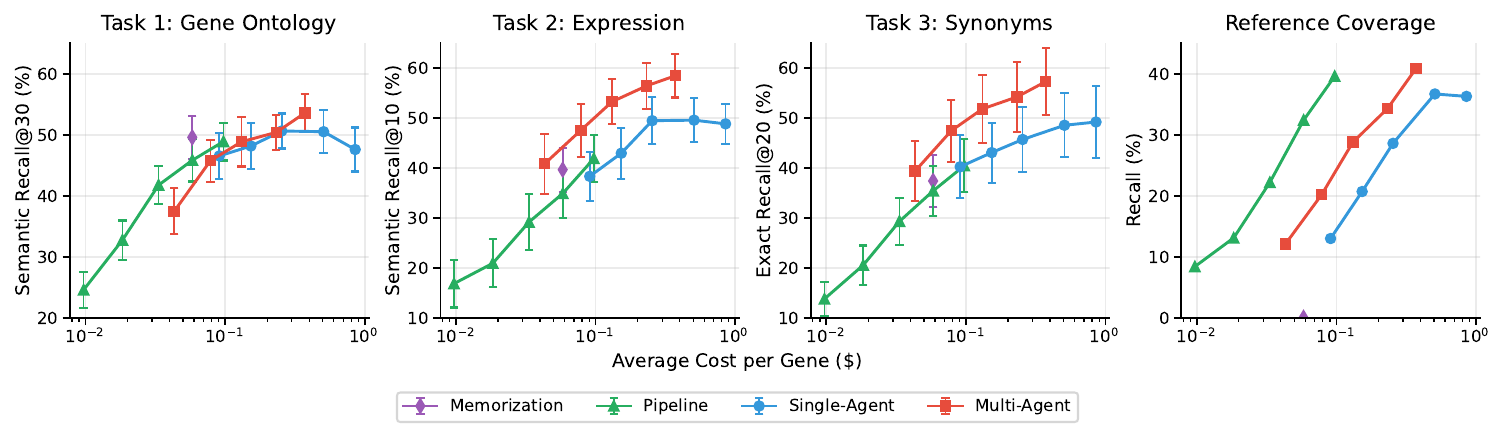}
    \caption{Cost-performance tradeoff across harnesses and paper budgets. The first three panels show primary task metrics with 95\% gene-bootstrap intervals; the fourth shows reference coverage, the fraction of unique curator-cited source papers for retained corpus-grounded labels retrieved.}
    \label{fig:pareto}
\end{figure*}

We compare harnesses across budgets of 1, 2, 4, 8, and 16 papers per gene, using GPT-5-mini as the backbone LLM (Figure~\ref{fig:pareto}). The budget is stated in the prompt and enforced in tool calls. Reference coverage (RefCov) is the fraction of unique curator-cited source papers for retained corpus-grounded labels that a run retrieves within its paper budget.

\noindent\textbf{Result 1: Retrieval improves over memorization in the Multi-Agent harness, with task-dependent gains.} At 16 papers, Multi-Agent achieves point estimates of 54\%, 58\%, and 57\% on \tgo{}, \texpr{}, and \tsyn{}, compared with 50\%, 40\%, and 37\% for Memorization. This corresponds to gains of 4, 18, and 20 percentage points, respectively. Gains are largest on \texpr{} and \tsyn{}, where organism-specific expression terms and historical names are less likely to appear frequently in model pretraining corpora.

\noindent\textbf{Result 2: Context management influences scaling.} Multi-Agent benefits most from larger paper budgets, improving from 37\% to 54\% on \tgo{}, 41\% to 58\% on \texpr{}, and 39\% to 57\% on \tsyn{} from 1 to 16 papers. Single-Agent peaks earlier on \tgo{} and \texpr{} and costs more at every budget, consistent with context dilution as raw paper text accumulates. This experiment does not isolate context dilution from retrieval mix or tool-use failures. The pattern nevertheless shows that harness design matters: partitioning context can improve recall while reducing cost relative to a monolithic agent.

\noindent\textbf{Result 3: Feedback loops improve over the fixed pipeline, especially beyond \tgo{}.} At matched budgets, Single-Agent and Multi-Agent have higher point estimates than Pipeline on \texpr{} and \tsyn{}, and Multi-Agent has the strongest overall average at 16 papers. Pipeline narrows the \tgo{} gap under the larger GO cutoff because it emits many more candidate GO annotations, but it cannot recover from an incorrect ontology mapping or expand search after discovering a synonym. Adaptive harnesses can revise queries and retry ontology resolution. Appendix~\ref{app:case-study} traces one Single-Agent run illustrating this search-read-refine loop.

\subsection{\texorpdfstring{Experiment 2: Diagnostic Failure Modes}{Experiment 2: Diagnostic Failure Modes}}
\label{sec:experiments:diagnostics}

Aggregate recall hides task- and harness-specific bottlenecks. We analyze Multi-Agent at 16 papers unless noted otherwise.

\noindent\textbf{GO aspect difficulty.} Task~1 spans Biological Process, Molecular Function, and Cellular Component (Section~\ref{sec:tasks:output}). Miss rates differ substantially by aspect: Molecular Function is easiest (34\% missed), Biological Process is hardest (50\% missed), and Cellular Component is intermediate (43\% missed). This pattern is consistent with how evidence appears in text: Molecular Function can be stated directly, while Biological Process often requires inference from experimental context.

\noindent\textbf{Context efficiency trades off with information fidelity.} Multi-Agent has higher aggregate point estimates than Single-Agent at 16 papers, but its advantage does not extend to every annotation. For each ground-truth annotation, we compute the best semantic similarity achieved by each method's top-$k$ predictions and determine a winner (Table~\ref{tab:win-draw-lose}). Multi-Agent wins roughly twice as often as Single-Agent on \tgo{} and \texpr{}, but Single-Agent still wins on 20\% and 26\% of annotations, respectively. These cases suggest that subagent delegation may lose details during compression, or that Single-Agent's accumulated context sometimes helps cross-paper synthesis.

\vspace{-10pt}
\begin{table}[ht]
\centering
\tablefont
\compacttablespacing
\caption{Per-annotation win/draw/lose between Single-Agent and Multi-Agent at 16 papers.}
\label{tab:win-draw-lose}
\begin{tabular}{lccc}
\toprule
\textbf{Task} & \textbf{Single Wins} & \textbf{Multi Wins} & \textbf{Draws} \\
\midrule
\taskgo{} & 184 (19.5\%) & 372 (39.4\%) & 389 (41.2\%) \\
\taskexpr{} & 86 (25.7\%) & 165 (49.4\%) & 83 (24.9\%) \\
\tasksyn{} & 37 (4.2\%) & 108 (12.1\%) & 745 (83.7\%) \\
\midrule
Total & 307 (14.2\%) & 645 (29.7\%) & 1217 (56.1\%) \\
\bottomrule
\end{tabular}
\end{table}

\flybench{} makes this tradeoff visible at the annotation level: aggregate recall favors context partitioning, but per-annotation wins reveal cases where compression may discard useful evidence. This points to richer intermediate representations from subagents, orchestrator-initiated follow-up queries, or hybrid harnesses that combine partitioned processing with selective full-context review.

\subsection{\texorpdfstring{Experiment 3: Model Sensitivity}{Experiment 3: Model Sensitivity}}
\label{sec:experiments:models}

We next hold the Multi-Agent harness fixed and compare model families at 16 papers per gene (Table~\ref{tab:model-comparison}). Appendix Table~\ref{tab:cross-family} reports the corresponding cross-family results for other harnesses.

\begin{table}[ht]
\centering
\tablefont
\compacttablespacing
\caption{Model comparison with the Multi-Agent harness at 16 papers/gene. Recall entries are percentages with 95\% gene-bootstrap half-widths; cost, tokens, and annotations are per-gene averages.}
\label{tab:model-comparison}
\begin{tabular}{lccccccc}
\toprule
\textbf{Model} & \textbf{Cost} & \textbf{Tokens} & \textbf{Pred. annot.} & \textcolor{taskblue}{\textbf{GO}} & \textcolor{taskgreen}{\textbf{Expr}} & \textcolor{taskpurple}{\textbf{Syn}} & \textbf{Avg.} \\
\midrule
GPT-5-mini & \$0.40 & 1.3M & 23 & 53.6{\scriptsize$\pm$3.1} & 58.5{\scriptsize$\pm$4.4} & \textbf{57.3{\scriptsize$\pm$6.6}} & 56.5{\scriptsize$\pm$3.4} \\
GPT-4o & \$2.74 & 1.2M & 17 & 47.6{\scriptsize$\pm$3.8} & 50.7{\scriptsize$\pm$4.8} & 48.6{\scriptsize$\pm$6.7} & 49.0{\scriptsize$\pm$3.6} \\
GPT-5 & \$5.89 & 1.4M & 29 & 57.1{\scriptsize$\pm$4.6} & \textbf{60.9{\scriptsize$\pm$7.4}} & 53.8{\scriptsize$\pm$8.0} & \textbf{57.3{\scriptsize$\pm$5.8}} \\
Claude Sonnet 4.6 & \$15.81 & 5.0M & 30 & \textbf{59.7{\scriptsize$\pm$5.7}} & 58.4{\scriptsize$\pm$5.7} & 46.4{\scriptsize$\pm$6.0} & 54.8{\scriptsize$\pm$4.2} \\
MiniMax M2.5 & \$0.62 & 2.0M & 26 & 53.1{\scriptsize$\pm$3.3} & 54.5{\scriptsize$\pm$4.6} & 53.0{\scriptsize$\pm$6.2} & 53.5{\scriptsize$\pm$3.0} \\
DeepSeek V3.2 & \$0.74 & 1.2M & 9 & 23.7{\scriptsize$\pm$5.3} & 23.3{\scriptsize$\pm$6.0} & 26.9{\scriptsize$\pm$5.9} & 24.7{\scriptsize$\pm$4.7} \\
\bottomrule
\end{tabular}
\vspace{-10pt}
\end{table}

Table~\ref{tab:model-comparison} shows task- and cost-sensitive model differences under the same Multi-Agent harness. GPT-5 and Claude Sonnet~4.6 have the highest point estimates on \texpr{} and \tgo{}, respectively, while GPT-5-mini has the highest \tsyn{} point estimate and comparable average recall at much lower cost. MiniMax~M2.5 is also competitive at low cost. DeepSeek V3.2 is substantially lower due to fewer submitted annotations and failed or empty outputs. Overall, \flybench{} distinguishes base-model capability, task fit, cost, annotation volume, and tool-use reliability in deployed agent configurations.

\subsection{\texorpdfstring{Experiment 4: Missing-Term Setting}{Experiment 4: Missing-Term Setting}}
\label{sec:experiments:missing}

We evaluate how agents handle concepts that lack ontology coverage under the \emph{missing-term setting of Task 1} (Section~\ref{sec:tasks:output}). The ontology search tool filters selected GO IDs, and the agent cannot output hidden GO identifiers; it must choose an available term or propose a natural-language description. Of 770 verified corpus-grounded (gene, GO term) pairs, 124 are selected to be \emph{missing terms}.

\noindent\textbf{Result: Agents find nearby ontology terms but rarely propose useful novel terms.}
Multi-Agent with GPT-5-mini reaches 48.4\% semantic recall@20 on missing terms versus 55.4\% on available terms (Appendix Table~\ref{tab:missing-term-comparison}). Much of this comes from approximate parent or sibling predictions, such as ``heart development'' for the hidden term ``adult heart development.'' The agent also produces 49 natural-language descriptions; an LLM-based resolver maps 44 to some GO term, but similarity to the hidden targets is low (mean 0.19; 6/44 exact matches). Agents can often identify the right biological neighborhood, but they rarely propose useful novel terms. Improving ontology-extension behavior remains an open challenge.

%%
%% 6. Discussion
%%

\vspace{-10pt}
\section{Discussion}
\vspace{-5pt}
\label{sec:discussion}

% \subsection{Limitations and Future Work}
\label{sec:discussion:limitations}

\noindent\textbf{Text-only evaluation.}
Scientific papers communicate through text and figures, yet \flybench{} evaluates only text extraction. Figures can contain critical evidence, such as microscopy expression patterns, that text may reference only indirectly. We therefore treat text-only baselines as a prerequisite for future multimodal evaluation with figures, tables, and full PDF layout.
% Our corpus-grounding labels identify papers containing evidence but not the modality of that evidence; some annotations labeled ``corpus-grounded'' may only be extractable from figures, making our text-based recall metrics conservative estimates of performance on genuinely text-extractable information.

\noindent\textbf{Corpus coverage.}
Our corpus comprises 16,898 PMC open-access papers. It covers 24\% of post-2000 FlyBase references (16,665 of 69,404), rising to 30--40\% after 2015 as open-access availability increases. This growing subset supports full reproducibility and increasingly represents recent literature. Broader coverage would require institutional partnerships or expanded PMC deposition.

\noindent\textbf{Ground truth completeness.}
FlyBase annotations are expert-curated and authoritative, but no biological database is exhaustive. Curators prioritize accuracy over coverage, and annotation lags behind the literature. Agents may therefore produce correct annotations absent from ground truth; measuring this requires human validation, which we leave to future work. For corpus grounding, we use automated recoverability verification plus domain-expert spot checks; future work could add a formal validation set to quantify agreement more precisely.

\noindent\textbf{Toward interactive curation.}
\flybench{} evaluates fully autonomous agents, but real-world deployment may favor human-agent collaboration. Future work could explore interactive settings where agents retrieve candidate annotations for curator review, request clarification on ambiguous cases, or learn from curator feedback.

\noindent\textbf{Ethics and broader impact.}
\label{sec:discussion:ethics}
\flybench{} is constructed from public FlyBase annotations (CC-BY 4.0) and PubMed Central open-access literature. Because benchmark outputs are candidate annotations, systems should be used to prioritize expert review rather than replace curator judgment; Appendix~\ref{app:responsible-impact} gives additional responsible-release details.

\vspace{-5pt}
\subsection{Conclusion}
\label{sec:discussion:conclusion}

We introduced \flybench{}, a benchmark for end-to-end agentic ontology curation from scientific literature. It turns a realistic curator workflow into a reproducible high-recall evaluation over 100 \textit{Drosophila} genes. Our baselines show that retrieval improves recall most clearly for expression and synonyms, and that these gains depend on context management, feedback loops, and tool-use reliability. By making these tradeoffs measurable, \flybench{} helps move scientific agents beyond answer finding toward dependable, literature-grounded knowledge base maintenance.

%%
%% Acknowledgments
%%
% \begin{acks}
% \todo{Add acknowledgments}
% \end{acks}

%%
%% Bibliography
%%
\bibliography{references}

%% Appendix
\clearpage
\appendix
\section*{Appendix Contents}
\label{app:contents}

\newcommand{\appendixtocitem}[2]{\noindent\hyperref[#1]{#2}\dotfill\pageref{#1}\par}

\appendixtocitem{app:checklist}{NeurIPS Paper Checklist}
\appendixtocitem{app:stats}{Dataset Details}
\appendixtocitem{app:data}{Data and Code Availability}
\appendixtocitem{app:responsible-impact}{Responsible AI and Broader Impact}
\appendixtocitem{app:repro}{Reproducibility Workflows}
\appendixtocitem{app:agent-benchmark-practice}{Uncertainty Reporting in Agent Benchmarks}
\appendixtocitem{app:prompts}{Prompt Design}
\appendixtocitem{app:metrics}{Evaluation Metrics Details}
\appendixtocitem{app:additional-experiments}{Additional Experiments}

\clearpage
\phantomsection
\section*{NeurIPS Paper Checklist}
\label{app:checklist}

\begin{enumerate}
    \item \textbf{Claims: Do the main claims made in the abstract and introduction accurately reflect the paper's contributions and scope?}
    \answerYes{} The abstract and introduction state the benchmark scope and contributions. Assumptions and limits are discussed in Sections~\ref{sec:tasks:eval}, \ref{sec:dataset}, and \ref{sec:discussion:limitations}.

    \item \textbf{Limitations: The authors are encouraged to create a separate ``Limitations'' section in their paper. The paper should point out any strong assumptions and how robust the results are to violations of these assumptions (e.g., independence assumptions, noiseless settings, model well-specification, asymptotic approximations only holding locally).}
    \answerYes{} Section~\ref{sec:discussion:limitations} discusses text-only evaluation, corpus coverage, ground-truth completeness, interactive curation, and system-dependent failure modes.

    \item \textbf{Theory, Assumptions and Proofs: If you are including theoretical results, did you state the full set of assumptions of all theoretical results, and did you include complete proofs of all theoretical results?}
    \answerNA{} The paper does not present theoretical results.

    \item \textbf{Experimental Result Reproducibility: If the contribution is a dataset or model, what steps did you take to make your results reproducible or verifiable?}
    \answerYes{} Appendix~\ref{app:repro} describes no-API table reproduction, API-dependent baseline reruns, metric policy, model settings, and evaluation commands.

    \item \textbf{Open Access to Data and Code: If you ran experiments, did you include the code, data, and instructions needed to reproduce the main experimental results (either in the supplemental material or as a URL)?}
    \answerYes{} Code and data URLs are provided in the abstract footnote and Appendix~\ref{app:data}; Appendix~\ref{app:repro} describes reproduction workflows.

    \item \textbf{Experimental Setting/ Details: If you ran experiments, did you specify all the training details (e.g., data splits, hyperparameters, how they were chosen)?}
    \answerYes{} Sections~\ref{sec:tasks:eval} and \ref{sec:experiments} specify the metrics, baselines, paper budgets, model settings needed to interpret the experiments; Appendix~\ref{app:repro} provides additional runnable details.

    \item \textbf{Experiment Statistical Significance: Does the paper report error bars suitably and correctly defined or other appropriate information about the statistical significance of the experiments?}
    \answerYes{} Figure~\ref{fig:pareto}, Table~\ref{tab:model-comparison}, and Appendix Table~\ref{tab:cross-family} report 95\% percentile bootstrap intervals for the primary baseline and model comparisons, computed by resampling genes with replacement and recomputing the same micro-averaged recall estimator. These intervals quantify variation across benchmark genes rather than stochastic provider reruns; secondary appendix ablations are descriptive frozen-run comparisons.

    \item \textbf{Experiments Compute Resource: For each experiment, does the paper provide sufficient information on the computer resources (type of compute workers, memory, time of execution) needed to reproduce the experiments?}
    \answerYes{} Appendix~\ref{app:repro} reports API costs, token usage, model providers, and runtime environment; Appendix~\ref{app:case-study} reports an example wall-clock run time. The main resource bottleneck is provider API cost rather than local CPU/GPU compute.

    \item \textbf{Code Of Ethics: Have you read the NeurIPS Code of Ethics and ensured that your research conforms to it?}
    \answerYes{} The work uses public scientific literature and public database annotations; Section~\ref{sec:discussion:ethics} and Appendix~\ref{app:responsible-impact} discuss responsible-use considerations.

    \item \textbf{Broader Impacts: If appropriate for the scope and focus of your paper, did you discuss potential negative societal impacts of your work?}
    \answerYes{} Section~\ref{sec:discussion:ethics} and Appendix~\ref{app:responsible-impact} discuss corpus biases and the risk of over-trusting automated curation without expert review.

    \item \textbf{Safeguards: Do you have safeguards in place for responsible release of models with a high risk for misuse (e.g., pretrained language models)?}
    \answerNA{} The paper releases a benchmark dataset, evaluation code, prompts, and baseline outputs rather than a high-risk pretrained model.

    \item \textbf{Licenses: If you are using existing assets (e.g., code, data, models), did you cite the creators and respect the license and terms of use?}
    \answerYes{} Appendix~\ref{app:data} documents FlyBase, ontology, and PMC-OA provenance and licenses, including per-paper PMC license metadata.

    \item \textbf{Assets: If you are releasing new assets, did you document them and provide these details alongside the assets?}
    \answerYes{} Appendix~\ref{app:data} documents the released dataset, code, provenance, license information, and Croissant metadata.

    \item \textbf{Crowdsourcing and Research with Human Subjects: If you used crowdsourcing or conducted research with human subjects, did you include the full text of instructions given to participants and screenshots, if applicable, as well as details about compensation (if any)?}
    \answerNA{} The benchmark does not use crowdsourcing or human-subject data collection.

    \item \textbf{IRB Approvals: Did you describe any potential participant risks and obtain Institutional Review Board (IRB) approvals (or an equivalent approval/review based on the requirements of your institution), if applicable?}
    \answerNA{} The work uses public literature and public database annotations, and does not involve human-subject research.

    \item \textbf{Declaration of LLM usage: Does the paper describe the usage of LLMs if it is an important, original, or non-standard component of the core methods in this research?}
    \answerYes{} LLMs are core experimental subjects in Section~\ref{sec:experiments}; Appendix~\ref{app:grounding-verification} also describes GPT-5 use for evidence-recoverability verification.
\end{enumerate}

\clearpage
%%
%% Appendix (unlimited pages)
%%
\section{Dataset Details}
\label{app:stats}

\begin{figure}[h!]
\centering
\includegraphics[width=\columnwidth]{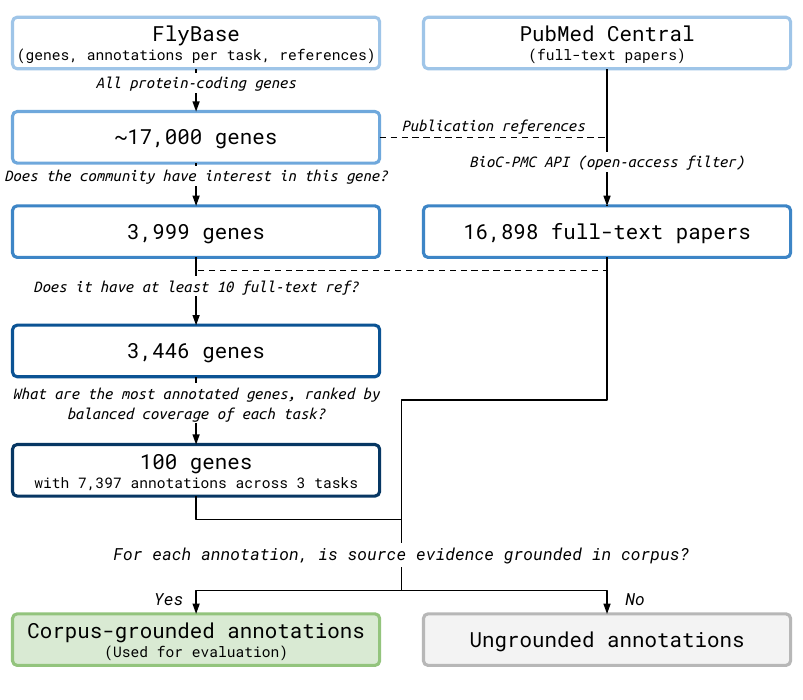}
\caption{Dataset construction pipeline. Genes are selected from FlyBase through a three-stage filter; the literature corpus is assembled from PubMed Central via the BioC-PMC API. Task~1 and Task~2 labels are corpus-grounded through evidence-recoverability verification over curator-linked source papers; Task~3 synonyms are grounded by text matching.}
\label{fig:data-construction}
\end{figure}

\begin{figure}[t!]
\centering
\includegraphics[width=\columnwidth]{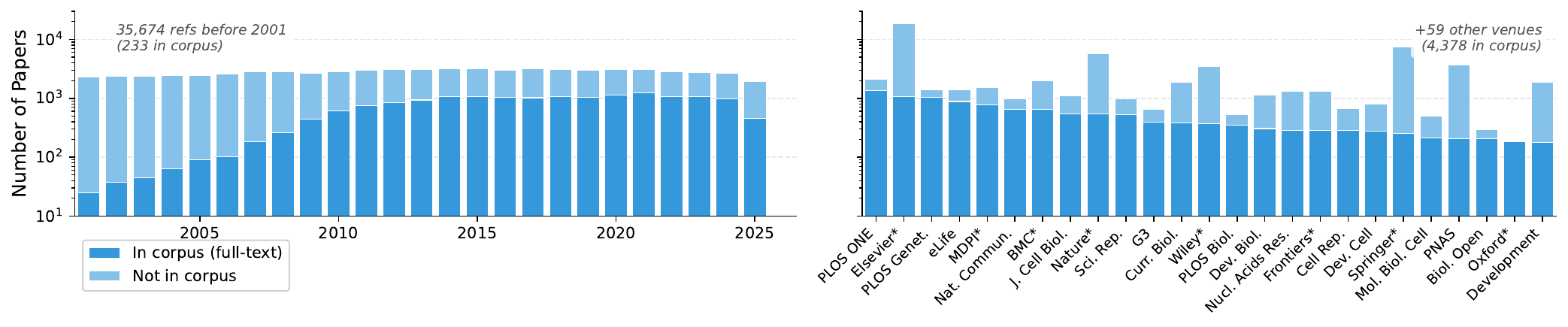}
\caption{Corpus composition by publication year (left) and venue (right). Darker bars: papers in our corpus; lighter bars: FlyBase references without open-access full text. *Publisher aggregates combining multiple journals.}
\label{fig:corpus-overview}
\end{figure}

\begin{table}[ht]
\centering
\tablefont
\compacttablespacing
\caption{Section coverage in the \flybench{} corpus.}
\label{tab:section-coverage}
\begin{tabular}{lrr}
\toprule
Section & Papers & Coverage \\
\midrule
Title & 16,898 & 100.0\% \\
Abstract & 16,783 & 99.3\% \\
Introduction & 16,556 & 98.0\% \\
Methods & 14,875 & 88.0\% \\
Results & 14,515 & 85.9\% \\
Discussion & 13,543 & 80.1\% \\
\bottomrule
\end{tabular}
\end{table}

\begin{table}[ht]
\centering
\tablefont
\compacttablespacing
\caption{Ontology coverage statistics from GO-basic and FlyBase ontology releases~\citep{GeneOntology_2023,FlyBase_2024}.}
\label{tab:ontology-stats}
\begin{tabular}{lrr}
\toprule
Ontology & Total Terms & Used in Benchmark \\
\midrule
Gene Ontology (GO-basic) & 48,196 & 1,228 \\
FlyBase Anatomy (FBbt) & 28,457 & 465 \\
FlyBase Development (FBdv) & 217 & 69 \\
\bottomrule
\end{tabular}
\end{table}

\begin{figure}[ht]
\centering
\scalebox{0.85}{% GO Hierarchy illustration using GO:0008049 (male courtship behavior)
% Shows tree-like structure with DAG nature (multiple parents)
\begin{tikzpicture}[
    goterm/.style={
        draw,
        rounded corners=2pt,
        font=\small,
        minimum height=1.4em,
        inner sep=4pt,
        fill=white,
        align=center
    },
    highlight/.style={goterm, fill=yellow!25, thick},
    arrow/.style={-stealth, thick, gray!60},
]

% Vertical spacing (reduced)
\def\vs{0.75cm}

% Top: ellipsis indicating root
\node[font=\footnotesize, text=gray] (root) at (0,0.15) {biological\_process};
\node[font=\footnotesize, text=gray] (dots) at (0,-0.15) {\vdots};

% Main hierarchy - symmetric layout
\node[goterm] (behavior) at (0, -0.5*\vs) {behavior {\footnotesize\texttt{GO:0007610}}};
\node[goterm] (repro) at (0, -1.6*\vs) {reproductive behavior {\footnotesize\texttt{GO:0019098}}};
\node[goterm] (mating) at (0, -2.7*\vs) {mating behavior {\footnotesize\texttt{GO:0007617}}};
\node[goterm] (male-mating) at (-2.1, -3.8*\vs) {male mating behavior {\footnotesize\texttt{GO:0060179}}};
\node[goterm] (courtship) at (2.1, -3.8*\vs) {courtship behavior {\footnotesize\texttt{GO:0007619}}};
\node[highlight] (target) at (0, -4.9*\vs) {\textbf{male courtship behavior} {\footnotesize\texttt{GO:0008049}}};

% Manhattan-style edges (orthogonal lines)
% Main chain: behavior -> repro -> mating
\draw[arrow] (behavior.south) -- (repro.north);
\draw[arrow] (repro.south) -- (mating.north);

% Mating -> male mating and courtship (fork)
\draw[arrow] (mating.south) -- ++(0,-0.12) -| (male-mating.north);
\draw[arrow] (mating.south) -- ++(0,-0.12) -| (courtship.north);

% Both parents -> target
\draw[arrow] (male-mating.south) -- ++(0,-0.12) -| ([xshift=-3mm]target.north);
\draw[arrow] (courtship.south) -- ++(0,-0.12) -| ([xshift=3mm]target.north);

\end{tikzpicture}}
\caption{Gene Ontology structure used for semantic scoring. GO forms a DAG; the highlighted term has two parents.}
\label{fig:go-hierarchy}
\end{figure}

\subsection{Evidence-Recoverability Verification}
\label{app:grounding-verification}

For Task~1 and Task~2, the initial corpus-grounding flag marked an annotation as in-corpus when at least one curator-linked source publication mapped to a paper in the released PMC corpus. This reference-level proxy can overestimate recoverability: the relevant evidence may appear only in figures, tables, supplements, scanned text, or other content absent from the released structured text.

We therefore ran evidence-recoverability verification using GPT-5. Each verification item consisted of (1) the target gene, (2) one FlyBase annotation rendered in natural language, including ontology identifiers and evidence metadata where available, and (3) the accessible structured text of one curator-linked paper, including title, abstract, and available section text. The verifier returned one of three verdicts---\texttt{supported}, \texttt{not\_supported}, or \texttt{uncertain}---plus a short rationale and an evidence span when supported. The prompt instructed the verifier to use moderate strictness: paraphrased or implicit evidence could count when a trained reader could reasonably derive the annotation, but tangential mentions or evidence absent from the provided text did not count.

When an annotation had multiple curator-linked source papers, we used OR aggregation: the annotation was considered corpus-grounded if any cited paper received a \texttt{supported} verdict. This makes the filter conservative with respect to agent observability while preserving labels supported by at least one accessible text source. Verification was used only to determine recoverability from the released corpus; it did not create new biological annotations.

The verification pass covered 1,529 annotation--paper pairs that were previously marked in-corpus by reference matching. After aggregation, 945 of 3,621 row-level \tgo{} annotations and 334 of 1,331 row-level \texpr{} annotations remained corpus-grounded. No completed judgment returned \texttt{uncertain}. Task~3 synonyms were not passed through this verifier because synonym support is lexical; they were grounded by text matching against the released corpus.

During benchmark development, domain experts qualitatively spot-checked verification rationales to assess whether supported labels were backed by appropriate evidence and rejected labels lacked recoverable textual support. This served as a reliability check on the automated grounding process rather than a separate formal validation set.

\section{Data and Code Availability}
\label{app:data}

\noindent\textbf{Code and Dataset.}
Code, prompts, and evaluation scripts are available in the public repository at \url{https://github.com/xingjian-zhang/flyaoc}. The combined dataset package is available at \url{https://huggingface.co/datasets/anonymous-042/flyaoc}. The dataset package contains the literature corpus, a merged per-gene benchmark file with gene metadata, verified ground-truth labels, and gene-to-paper mappings, ontology source files, hidden-term benchmark inputs, per-paper license/provenance metadata, and Croissant metadata. Model predictions and evaluated result dumps are not part of the dataset package; they are reproducibility artifacts in the code release.

\noindent\textbf{Literature Corpus.}
The corpus comprises 16,898 full-text papers from the PubMed Central Open Access (PMC-OA) subset, retrieved via the BioC-PMC API---an approved bulk retrieval method.\footnote{\url{https://pmc.ncbi.nlm.nih.gov/tools/openftlist/}} PMC-OA articles carry varying licenses (including CC-BY, CC-BY-NC, CC-BY-NC-SA, CC-BY-NC-ND, CC0, author manuscript, and other terms). We provide per-paper license/provenance metadata and PMCID manifests so users can inspect article-level terms or re-retrieve source articles from PMC when stricter license controls are required.

\noindent\textbf{Ground Truth Annotations.}
Annotations are derived from FlyBase release FB2025\_04~\citep{FlyBase_2024}. FlyBase data is released under CC-BY 4.0, permitting use, redistribution, and derivative works with attribution.\footnote{\url{https://wiki.flybase.org/wiki/FlyBase:About}} Source files:
\begin{itemize}
    \item Task 1: \texttt{gene\_association.fb} (GAF 2.2 format)
    \item Task 2: \texttt{curated\_expression\_fb\_2025\_04.tsv}
    \item Task 3: \texttt{fb\_synonym\_fb\_2025\_04.tsv}
\end{itemize}
We release processed ground truth with corpus-grounding labels indicating which annotations are derivable from our corpus.

\noindent\textbf{Ontologies.}
Gene Ontology (GO-basic) is released under CC-BY 4.0~\citep{GeneOntology_2023}. FlyBase ontologies (FBbt, FBdv) are released under CC-BY 4.0 as part of the OBO Foundry~\citep{FlyBase_2024}. We include the exact OBO source files used by the benchmark; searchable indices can be regenerated from the code release.

\noindent\textbf{Croissant Metadata.}
The dataset release includes \texttt{croissant.json} with core fields and minimal Responsible AI fields for limitations, biases, sensitive information, intended use, social impact, synthetic-data status, source datasets, and provenance activities; per-paper source licenses are separately documented in \texttt{pmc\_license\_manifest.jsonl}.

\section{Responsible AI and Broader Impact}
\label{app:responsible-impact}

\noindent\textbf{Data provenance and licensing.}
\flybench{} uses public FlyBase annotations and ontology resources released under CC-BY 4.0, together with article text from the PubMed Central Open Access subset. PMC-OA articles carry heterogeneous article-level licenses, so the release includes per-paper license and provenance metadata to support downstream audits and re-retrieval from PMC when stricter license controls are required. The benchmark does not use private laboratory records, reviewer comments, crowdsourcing data, or newly collected human-subject data.

\noindent\textbf{Limitations and biases.}
The corpus reflects the scientific record and the open-access subset available through PMC. It is biased toward English-language publications, venues and years with open full text, well-studied genes, and findings that have already been published and curated by FlyBase. It also underrepresents evidence that appears only in figures, tables, supplements, or non-open-access articles. Benchmark performance should therefore be interpreted as performance on reproducible, text-accessible curation tasks, not as exhaustive coverage of \textit{Drosophila} biology or all evidence available to expert curators.

\noindent\textbf{Intended use.}
The dataset is intended for evaluating methods that retrieve literature, extract candidate biological annotations, normalize them to controlled vocabularies, and report reproducible recall against verified corpus-grounded labels. It is suitable for comparing agent harnesses, retrieval strategies, context-management methods, and ontology-resolution procedures. It is not intended to support automatic database updates without expert review, clinical or agricultural decision-making, or claims that a predicted annotation is biologically correct without checking the cited evidence.

\noindent\textbf{Potential harms and mitigations.}
The main risks are over-trusting incorrect candidate annotations, reinforcing biases in the literature and curated databases, or treating retrieval success as a substitute for biological validation. We mitigate these risks by releasing source-paper provenance, article-level license metadata, ontology-normalized labels, frozen prediction artifacts, and no-API reproduction workflows. The benchmark releases evaluation data, prompts, code, and baseline outputs rather than a deployed curation service or high-risk pretrained model. We recommend using system outputs to prioritize expert curator review and evidence inspection, not to replace curator judgment.

\section{Reproducibility Workflows}
\label{app:repro}

\noindent\textbf{No-API table reproduction.}
The released code separates exact table reproduction from API-dependent reruns. Users can reproduce the reported tables without model-provider credentials by running \texttt{uv sync}, \texttt{uv run python scripts/smoke\_test.py}, and \texttt{uv run python scripts/reproduce\_tables.py}. This workflow loads benchmark labels and ontology files from the Hugging Face dataset, evaluates the frozen normalized predictions in \texttt{artifacts/predictions/}, and writes regenerated table files under \texttt{artifacts/tables/}. These frozen predictions are the source for exact paper-number reproduction.

\noindent\textbf{API baseline reruns.}
Users who want to rerun the four main baselines can install the optional baseline dependencies with \texttt{uv sync --extra baselines}, configure either \texttt{OPENAI\_API\_KEY} for direct OpenAI access or \texttt{OPENAI\_API\_KEY} plus \texttt{OPENAI\_BASE\_URL} for an OpenAI-compatible proxy, and run \texttt{flyaoc-run-baseline}. Each run writes raw per-gene outputs, a normalized \texttt{predictions.jsonl}, and a \texttt{run\_summary.json}; the generated predictions can be evaluated with \texttt{flyaoc-evaluate-predictions}. API reruns reproduce the experimental procedure under current provider behavior, while frozen predictions reproduce the exact reported tables.

\noindent\textbf{New method evaluation.}
New methods do not need to use our baseline harness. They can write normalized prediction JSONL with gene identifiers, run status, ranked Task~1/2 predictions, and Task~3 synonym lists, then call \texttt{flyaoc-evaluate-predictions}. This supports methods that use different prompts, retrieval systems, model providers, or task decompositions while preserving a common evaluator.

\noindent\textbf{Metric policy.}
The canonical paper metrics are micro-averaged over deduplicated verified corpus-grounded facts: Task~1 uses GO semantic recall@30, Task~2 uses anatomy semantic recall@10, and Task~3 uses synonym exact recall@20 (Section~\ref{sec:tasks:eval}). The evaluator loads denominators from \texttt{benchmark.jsonl}: 770 GO term annotations, 252 expression anatomy-term annotations, and 457 synonym strings. Developmental-stage IDs remain in normalized Task~2 outputs, but are not included in the canonical primary metrics reported here. Failed or empty-output runs are evaluated as empty predictions, contributing zero numerator while preserving their verified-fact denominators. Macro averages and additional recall cutoffs are emitted for diagnostics and appendix tables.

\noindent\textbf{Models and hyperparameters.}
Primary results use GPT-5-mini (Feb 2026) via OpenAI API. Model comparison (Section~\ref{sec:experiments:models}) additionally evaluates GPT-4o and GPT-5 via OpenAI, and Claude Sonnet~4.6, MiniMax M2.5, and DeepSeek V3.2 via Amazon Bedrock. The frozen prediction artifacts use identifiers \texttt{openai\_gpt-5-mini}, \texttt{openai\_gpt-4o}, \texttt{openai\_gpt-5}, \texttt{bedrock\_claude-sonnet-4}, \texttt{bedrock\_minimax-m2.5}, and \texttt{bedrock\_deepseek-v3.2}. API reruns reproduce the procedure under the current provider snapshot for the configured identifier. Temperature is set to 1.0 for all models; other parameters use provider defaults. Agent execution is capped at 50 turns per gene with a cost limit of \$1--10 depending on model.

\noindent\textbf{Costs, environment, and code structure.}
Table~\ref{tab:full-cost} reports total API costs. At 16 papers per gene, Multi-Agent costs roughly \$0.4/gene, Single-Agent \$0.85/gene, and Pipeline \$0.10/gene. The code targets Python 3.11+. The ported original paper harness lives under \texttt{src/agent/}, while \texttt{src/flyaoc/baselines/} provides public adapters, command-line entry points, and output normalization. The missing-term experiment requires a separate hidden-label setup and is not part of the public v1 rerun harness.

\section{Uncertainty Reporting in Agent Benchmarks}
\label{app:agent-benchmark-practice}

Agent benchmarks are often expensive to run because each task can require long tool-using trajectories, external environments, and paid model APIs. As a result, several influential benchmarks report controlled point estimates without formal confidence intervals or significance tests, including SWE-bench~\citep{jimenez2024swebench}, WebArena~\citep{zhou2024webarena}, GAIA~\citep{mialon2024gaia}, AgentBench~\citep{liu2024agentbench}, and OSWorld~\citep{xie2024osworld}. Their narratives typically emphasize task-scoped success rates, large capability gaps, and qualitative failure modes rather than formal pairwise superiority.

Recent work also shows two useful directions for more careful reporting. $\tau$-bench introduces \textit{pass}$^k$ to measure whether stochastic tool-using agents succeed consistently across repeated trials~\citep{yao2024taubench}; this captures provider/run variance but requires multiple executions per task. BrowserGym reports uncertainty for benchmark success rates and explicitly documents budget-driven omissions in a large multi-benchmark web-agent evaluation~\citep{lesellier2025browsergym}.

Our reporting follows this practice while adding a conservative uncertainty estimate where it is affordable from frozen artifacts. For primary comparison surfaces (Figure~\ref{fig:pareto}, Table~\ref{tab:model-comparison}, and Table~\ref{tab:cross-family}), we bootstrap over genes, the benchmark examples, and recompute the same micro-averaged recall estimator. These intervals quantify sensitivity to the sampled gene set and the dependence among annotations within a gene. They should not be interpreted as repeated API-run intervals. Accordingly, close model or harness differences are described as point-estimate patterns, while large gaps, cost tradeoffs, and failure modes are interpreted as benchmark observations under fixed released runs.

\section{Prompt Design}
\label{app:prompts}

All prompts are available in our code release. Here we summarize their key components.

\noindent\textbf{Task Prompt.}
Specifies the gene to annotate (symbol, ID, summary), enforces the paper budget, lists available tools, and provides strategy tips. Single-Agent references \texttt{get\_paper\_text}; Multi-Agent references \texttt{analyze\_papers\_batch}.

\noindent\textbf{System Prompt.}
Shared by Single-Agent and Multi-Agent. Defines the three extraction tasks, explains GO annotation basics (aspects, qualifiers), provides textual patterns for identifying valid evidence (e.g., ``X mutants fail to...'' for biological process), lists what to avoid (speculation, background statements, homology claims), and specifies the JSON output schema.

\noindent\textbf{Multi-Agent Addendum.}
Instructs the orchestrator to delegate paper reading to subagents via \texttt{analyze\_papers\_batch}, which returns pre-resolved annotations with ontology IDs. Describes the aggregation role: deduplicate, resolve conflicts, filter noise, combine evidence across papers.

\noindent\textbf{Memorization Prompt.}
Explicitly states no corpus access. Instructs the model to rely on parametric knowledge, use ontology search tools to find correct IDs, and be conservative. Evidence format uses \texttt{pmcid: null} with explanatory text.

\noindent\textbf{Pipeline Extraction Prompt.}
Used by the Pipeline baseline for per-paper extraction. Instructs the model to output natural language descriptions (not ontology IDs) for functions, anatomy, and stages. A separate resolution step maps descriptions to IDs post-hoc.

\section{Evaluation Metrics Details}
\label{app:metrics}

\subsection{Aggregation}

To aggregate across the 100 benchmark genes, the primary metric is \textbf{micro-averaged} recall:
\[
\frac{\sum_i \text{numerator}_i}{\sum_i |\text{GT}_i|}
\]
where each annotation contributes equally regardless of gene. This weights well-studied genes with many annotations proportionally more, reflecting cases where automated curation would save the most effort. We additionally report \textbf{macro-averaged} recall:
\[
\frac{1}{N}\sum_i \text{Recall}_i@k
\]
which gives equal weight per gene and checks that conclusions are consistent across aggregation choices. Failed or empty-output runs are scored as zero recall for that gene rather than excluded.

\subsection{Secondary Metrics}

Beyond the primary corpus-grounded semantic recall@$k$ reported in the main text, we define additional diagnostic metrics and report precision/F1 at the primary cutoffs.

\noindent\textbf{Full-reference Semantic Recall@$k$.}
Evaluates against all ground-truth annotations, including those from sources outside our corpus. This measures end-to-end task difficulty but conflates extraction capability with corpus coverage limitations.

\noindent\textbf{Semantic Precision and F1.}
For completeness, we report precision at the task-specific primary cutoff using the same similarity-weighted approach:
\[
\text{Semantic Precision@}k = \frac{\sum_{p \in \text{Pred}_{1:k}} \max_{g \in \text{GT}} \text{sim}(p, g)}{|\text{Pred}_{1:k}|}
\]
Semantic F1 is the harmonic mean of semantic precision and recall. We de-emphasize precision because the benchmark is designed around high-recall candidate generation: missed annotations force curators back into open-ended literature search, while over-predictions can be filtered once evidence is surfaced. Ground truth incompleteness also makes false positive counts unreliable.

Table~\ref{tab:precision-f1} reports semantic precision, recall, and F1 at the primary cutoffs for all methods at 16 papers per gene (except Memorization at 0 papers). Precision varies substantially across methods: Memorization achieves the strongest synonym precision, while Multi-Agent trades some GO and expression precision for higher recall. Pipeline shows the lowest precision (20--29\%), reflecting its tendency to produce many candidate annotations that require filtering.

\begin{table}[ht]
\centering
\densetablefont
\densetablespacing
\caption{Semantic precision, recall, and F1 (micro-averaged, corpus-grounded) at the primary cutoffs: GO@$30$, Expr@$10$, and Syn@$20$. Memorization uses 0 papers. Syn uses exact matching.}
\label{tab:precision-f1}
\begin{tabular}{lcccccccccc}
\toprule
& \multicolumn{3}{c}{\textbf{\tgo{}}} & \multicolumn{3}{c}{\textbf{\texpr{}}} & \multicolumn{3}{c}{\textbf{\tsyn{}}} \\
\cmidrule(lr){2-4} \cmidrule(lr){5-7} \cmidrule(lr){8-10}
\textbf{Method} & P & R & F1 & P & R & F1 & P & R & F1 \\
\midrule
Memorization & 48.5 & 49.6 & 49.1 & 36.5 & 39.7 & 38.1 & 62.0 & 37.4 & 46.7 \\
Pipeline & 26.4 & 48.9 & 34.3 & 28.8 & 41.9 & 34.1 & 20.3 & 40.5 & 27.0 \\
Single-Agent & 47.4 & 47.6 & 47.5 & 36.1 & 48.9 & 41.5 & 48.7 & 49.2 & 49.0 \\
Multi-Agent & 44.0 & 53.6 & 48.3 & 33.9 & 58.5 & 42.9 & 47.8 & 57.3 & 52.1 \\
\bottomrule
\end{tabular}
\end{table}

\clearpage
\subsection{Full Results Tables}

Tables~\ref{tab:full-go}--\ref{tab:full-cost} provide complete recall@$k$ results across all paper budgets. Each task table reports micro-averaged corpus-grounded recall at multiple $k$ values and macro-averaged corpus-grounded recall at the primary $k$. The body text reports micro-averaged semantic recall@30 for \tgo{}, anatomy semantic recall@10 for \texpr{}, and exact recall@20 for \tsyn{}. Showing additional $k$ values reveals how the ranking cutoff interacts with each harness---agentic methods produce fewer than $k$ predictions at higher cutoffs, while Pipeline continues to benefit. Full results at all $k$ values and evaluation scopes are available in our data release.

\begin{table}[ht]
\centering
\densetablefont
\densetablespacing
\caption{Task 1 (\taskgo{}): Semantic recall@$k$ scaling (corpus-grounded). Bold: best at primary cutoff ($k{=}30$).}
\label{tab:full-go}
\begin{tabular}{llcccccc}
\toprule
& \textbf{Method} & \textbf{0p} & \textbf{1p} & \textbf{2p} & \textbf{4p} & \textbf{8p} & \textbf{16p} \\
\midrule
\multicolumn{8}{l}{\textit{Micro-averaged}} \\
\midrule
\multirow{4}{*}{$k{=}5$}
& Memorization & 39.4 & -- & -- & -- & -- & -- \\
& Single-Agent & -- & 41.9 & 40.7 & 39.9 & 39.4 & 39.3 \\
& Multi-Agent & -- & 34.7 & 38.0 & 37.5 & 37.3 & 40.3 \\
& Pipeline & -- & 20.9 & 23.0 & 23.4 & 23.8 & 23.1 \\
\midrule
\multirow{4}{*}{$k{=}10$}
& Memorization & 49.1 & -- & -- & -- & -- & -- \\
& Single-Agent & -- & 46.5 & 48.1 & 50.4 & 50.5 & 47.6 \\
& Multi-Agent & -- & 37.5 & 45.8 & 47.7 & 49.7 & 52.7 \\
& Pipeline & -- & 24.1 & 31.1 & 32.6 & 33.2 & 33.6 \\
\midrule
\multirow{4}{*}{$k{=}20$}
& Memorization & 49.6 & -- & -- & -- & -- & -- \\
& Single-Agent & -- & 46.5 & 48.2 & 50.6 & 50.5 & 47.6 \\
& Multi-Agent & -- & 37.5 & 45.8 & 48.9 & 50.4 & 53.6 \\
& Pipeline & -- & 24.6 & 32.7 & 41.1 & 43.2 & 44.0 \\
\midrule
\multirow{4}{*}{$k{=}30$}
& Memorization & 49.6 & -- & -- & -- & -- & -- \\
& Single-Agent & -- & \textbf{46.5} & \textbf{48.2} & \textbf{50.6} & \textbf{50.5} & 47.6 \\
& Multi-Agent & -- & 37.5 & 45.8 & 48.9 & 50.4 & \textbf{53.6} \\
& Pipeline & -- & 24.6 & 32.7 & 41.8 & 45.8 & 48.9 \\
\midrule
\multicolumn{8}{l}{\textit{Macro-averaged}} \\
\midrule
\multirow{4}{*}{$k{=}30$}
& Memorization & 51.2 & -- & -- & -- & -- & -- \\
& Single-Agent & -- & \textbf{49.5} & \textbf{52.5} & \textbf{53.5} & \textbf{54.5} & 50.9 \\
& Multi-Agent & -- & 41.7 & 49.6 & 53.1 & 53.4 & \textbf{56.3} \\
& Pipeline & -- & 22.5 & 31.5 & 41.3 & 44.9 & 49.2 \\
\bottomrule
\end{tabular}
\end{table}

\begin{table}[ht]
\centering
\densetablefont
\densetablespacing
\caption{Task 2 (\taskexpr{}): Anatomy semantic recall@$k$ scaling (corpus-grounded). Bold: best at primary cutoff ($k{=}10$).}
\label{tab:full-expr}
\begin{tabular}{llcccccc}
\toprule
& \textbf{Method} & \textbf{0p} & \textbf{1p} & \textbf{2p} & \textbf{4p} & \textbf{8p} & \textbf{16p} \\
\midrule
\multicolumn{8}{l}{\textit{Micro-averaged}} \\
\midrule
\multirow{4}{*}{$k{=}3$}
& Memorization & 39.2 & -- & -- & -- & -- & -- \\
& Single-Agent & -- & 37.8 & 41.4 & 45.0 & 43.9 & 45.8 \\
& Multi-Agent & -- & 35.8 & 41.7 & 47.0 & 48.5 & 48.3 \\
& Pipeline & -- & 16.9 & 20.8 & 27.2 & 31.9 & 34.8 \\
\midrule
\multirow{4}{*}{$k{=}5$}
& Memorization & 39.7 & -- & -- & -- & -- & -- \\
& Single-Agent & -- & 38.3 & 43.0 & 49.5 & 49.5 & 48.8 \\
& Multi-Agent & -- & 39.0 & 47.0 & 51.5 & 54.4 & 55.2 \\
& Pipeline & -- & 16.9 & 20.9 & 29.0 & 34.3 & 37.8 \\
\midrule
\multirow{4}{*}{$k{=}10$}
& Memorization & 39.7 & -- & -- & -- & -- & -- \\
& Single-Agent & -- & 38.3 & 43.0 & 49.5 & 49.6 & 48.9 \\
& Multi-Agent & -- & \textbf{40.9} & \textbf{47.6} & \textbf{53.3} & \textbf{56.5} & \textbf{58.5} \\
& Pipeline & -- & 16.9 & 20.9 & 29.2 & 34.9 & 41.9 \\
\midrule
\multicolumn{8}{l}{\textit{Macro-averaged}} \\
\midrule
\multirow{4}{*}{$k{=}10$}
& Memorization & 35.8 & -- & -- & -- & -- & -- \\
& Single-Agent & -- & 34.6 & 38.6 & 45.1 & 43.7 & 44.0 \\
& Multi-Agent & -- & \textbf{35.2} & \textbf{41.6} & \textbf{48.4} & \textbf{51.0} & \textbf{51.2} \\
& Pipeline & -- & 13.7 & 18.1 & 23.0 & 29.3 & 35.3 \\
\bottomrule
\end{tabular}
\end{table}

\begin{table}[ht]
\centering
\densetablefont
\densetablespacing
\caption{Task 3 (\tasksyn{}): Exact recall@$k$ scaling (corpus-grounded). Bold: best at primary cutoff ($k{=}20$).}
\label{tab:full-syn}
\begin{tabular}{llcccccc}
\toprule
& \textbf{Method} & \textbf{0p} & \textbf{1p} & \textbf{2p} & \textbf{4p} & \textbf{8p} & \textbf{16p} \\
\midrule
\multicolumn{8}{l}{\textit{Micro-averaged}} \\
\midrule
\multirow{4}{*}{$k{=}3$}
& Memorization & 33.3 & -- & -- & -- & -- & -- \\
& Single-Agent & -- & 35.9 & 35.2 & 32.8 & 32.4 & 29.1 \\
& Multi-Agent & -- & 35.9 & 35.0 & 35.2 & 31.9 & 30.4 \\
& Pipeline & -- & 12.7 & 17.7 & 18.6 & 19.0 & 19.7 \\
\midrule
\multirow{4}{*}{$k{=}5$}
& Memorization & 37.4 & -- & -- & -- & -- & -- \\
& Single-Agent & -- & 39.8 & 42.9 & 44.0 & 47.3 & 45.3 \\
& Multi-Agent & -- & 38.9 & 46.0 & 49.9 & 48.1 & 50.8 \\
& Pipeline & -- & 13.1 & 19.3 & 25.2 & 25.6 & 27.1 \\
\midrule
\multirow{4}{*}{$k{=}10$}
& Memorization & 37.4 & -- & -- & -- & -- & -- \\
& Single-Agent & -- & 40.3 & 43.1 & 45.7 & 48.6 & 49.2 \\
& Multi-Agent & -- & 39.4 & 47.5 & 51.9 & 54.3 & 57.3 \\
& Pipeline & -- & 13.8 & 20.6 & 28.4 & 32.6 & 34.8 \\
\midrule
\multirow{4}{*}{$k{=}20$}
& Memorization & 37.4 & -- & -- & -- & -- & -- \\
& Single-Agent & -- & \textbf{40.3} & 43.1 & 45.7 & 48.6 & 49.2 \\
& Multi-Agent & -- & 39.4 & \textbf{47.5} & \textbf{51.9} & \textbf{54.3} & \textbf{57.3} \\
& Pipeline & -- & 13.8 & 20.6 & 29.3 & 35.4 & 40.5 \\
\midrule
\multicolumn{8}{l}{\textit{Macro-averaged}} \\
\midrule
\multirow{4}{*}{$k{=}20$}
& Memorization & 45.6 & -- & -- & -- & -- & -- \\
& Single-Agent & -- & \textbf{48.4} & 53.2 & 55.8 & 60.3 & 60.3 \\
& Multi-Agent & -- & 48.1 & \textbf{57.6} & \textbf{63.5} & \textbf{64.8} & \textbf{68.1} \\
& Pipeline & -- & 15.2 & 22.7 & 31.5 & 38.0 & 42.0 \\
\bottomrule
\end{tabular}
\end{table}

\begin{table}[ht]
\centering
\densetablefont
\densetablespacing
\caption{Reference coverage and cost. RefCov: unique curator-cited source papers for retained corpus-grounded labels retrieved (\%). Cost: total API cost for 100 genes (USD, GPT-5-mini).}
\label{tab:full-cost}
\begin{tabular}{llcccccc}
\toprule
\textbf{Metric} & \textbf{Method} & \textbf{0p} & \textbf{1p} & \textbf{2p} & \textbf{4p} & \textbf{8p} & \textbf{16p} \\
\midrule
\multirow{4}{*}{RefCov}
& Memorization & 0.0 & -- & -- & -- & -- & -- \\
& Single-Agent & -- & 13.0 & 20.7 & 28.6 & \textbf{36.7} & 36.3 \\
& Multi-Agent & -- & 12.1 & 20.2 & \textbf{28.8} & 34.3 & \textbf{40.9} \\
& Pipeline & -- & 8.4 & 13.1 & 22.2 & 32.4 & 39.6 \\
\midrule
\multirow{4}{*}{Cost (\$)}
& Memorization & 5.82 & -- & -- & -- & -- & -- \\
& Single-Agent & -- & 9.09 & 15.29 & 25.45 & 50.74 & 85.35 \\
& Multi-Agent & -- & 4.32 & 7.85 & 13.13 & 23.23 & 37.25 \\
& Pipeline & -- & 0.97 & 1.84 & 3.34 & 5.83 & 9.73 \\
\bottomrule
\end{tabular}
\end{table}

\clearpage

% New experiments added during rebuttal revision
\section{Additional Experiments}
\label{app:additional-experiments}

\subsection{Missing-Term Setting}
\label{app:missing-term}

The missing-term setting tests whether agents can respond when a relevant biological concept lacks visible ontology support. We hide rare GO leaf terms that appear in only one corpus paper. The ontology search tool filters selected GO IDs, and agents cannot output hidden identifiers; they must choose an available term or propose a natural-language description. Of 770 verified corpus-grounded (gene, GO term) pairs, 124 are selected as missing terms.

We run Multi-Agent with GPT-5-mini at 16 papers per gene. The agent produced 49 natural-language descriptions across 100 genes. We then use an LLM-based resolver, which searches the full ontology including hidden terms, to map descriptions back to GO terms. Of 49 descriptions, 44 (90\%) resolve to some GO term, but similarity to the hidden targets is low (mean 0.19; 6/44 exact matches). This hidden-term analysis uses its original $k{=}20$ evaluation because it requires a separate hidden-label setup and is not part of the public no-API table reproduction path.

\begin{table}[ht]
\centering
\tablefont
\compacttablespacing
\caption{Statistics on available- vs.\ missing-term groups.}
\label{tab:missing-term-comparison}
\begin{tabular}{lcc}
\toprule
& \textbf{Available-Term} & \textbf{Missing-Term} \\
\midrule
Verified GO gene-term pairs & 646 & 124 \\
Semantic recall@20 & 55.4\% & 48.4\% \\
\bottomrule
\end{tabular}
\end{table}

\subsection{Cross-Family Harness Comparison}
\label{app:cross-family}

\begin{table}[ht]
\centering
\footnotesize
\setlength{\tabcolsep}{2.4pt}\renewcommand{\arraystretch}{0.95}
\caption{Cross-family model and harness comparison at budget 16. Recall entries are percentages with smaller-font 95\% gene-bootstrap half-widths.}
\label{tab:cross-family}
\begin{tabular}{lccccccccc}
\toprule
& \multicolumn{3}{c}{\textbf{Memorization}} & \multicolumn{3}{c}{\textbf{Single-Agent}} & \multicolumn{3}{c}{\textbf{Multi-Agent}} \\
\cmidrule(lr){2-4} \cmidrule(lr){5-7} \cmidrule(lr){8-10}
\textbf{Model} & \textcolor{taskblue}{\textbf{GO}} & \textcolor{taskgreen}{\textbf{Expr}} & \textcolor{taskpurple}{\textbf{Syn}} & \textcolor{taskblue}{\textbf{GO}} & \textcolor{taskgreen}{\textbf{Expr}} & \textcolor{taskpurple}{\textbf{Syn}} & \textcolor{taskblue}{\textbf{GO}} & \textcolor{taskgreen}{\textbf{Expr}} & \textcolor{taskpurple}{\textbf{Syn}} \\
\midrule
Claude Sonnet 4.6 & \textbf{52.7{\scriptsize$\pm$2.9}} & \textbf{44.4{\scriptsize$\pm$4.4}} & \textbf{38.1{\scriptsize$\pm$5.3}} & \textbf{58.9{\scriptsize$\pm$3.2}} & \textbf{53.9{\scriptsize$\pm$6.0}} & 48.8{\scriptsize$\pm$6.0} & \textbf{59.7{\scriptsize$\pm$5.6}} & 58.4{\scriptsize$\pm$5.7} & 46.4{\scriptsize$\pm$6.0} \\
GPT-5-mini & 49.6{\scriptsize$\pm$3.4} & 39.7{\scriptsize$\pm$4.4} & 37.4{\scriptsize$\pm$5.2} & 47.6{\scriptsize$\pm$3.5} & 48.9{\scriptsize$\pm$3.9} & \textbf{49.2{\scriptsize$\pm$7.0}} & 53.6{\scriptsize$\pm$3.2} & \textbf{58.5{\scriptsize$\pm$4.4}} & \textbf{57.3{\scriptsize$\pm$6.8}} \\
MiniMax M2.5 & 46.9{\scriptsize$\pm$3.5} & 40.5{\scriptsize$\pm$5.3} & 26.9{\scriptsize$\pm$4.3} & 50.3{\scriptsize$\pm$2.9} & 46.6{\scriptsize$\pm$4.7} & 43.3{\scriptsize$\pm$6.4} & 53.1{\scriptsize$\pm$3.3} & 54.5{\scriptsize$\pm$4.7} & 53.0{\scriptsize$\pm$6.3} \\
DeepSeek V3.2$^\dagger$ & 29.0{\scriptsize$\pm$5.6} & 26.4{\scriptsize$\pm$5.5} & 16.0{\scriptsize$\pm$4.0} & 3.5{\scriptsize$\pm$4.2} & 2.6{\scriptsize$\pm$3.2} & 1.8{\scriptsize$\pm$2.0} & 23.7{\scriptsize$\pm$5.1} & 23.3{\scriptsize$\pm$6.1} & 26.9{\scriptsize$\pm$5.9} \\
\bottomrule
\end{tabular}
\tablenote{$^\dagger$DeepSeek V3.2 returned failed or empty outputs on 36/100 (Memo), 94/100 (Single), and 40/100 (Multi) genes due to context overflow and tool-call errors; these runs are scored as zero recall. Bold marks highest point estimate within each harness/task column.}
\end{table}

The best point-estimate harness depends on model family and task: Multi-Agent is strongest for GPT-5-mini and MiniMax on most tasks, while Claude favors Multi-Agent on \tgo{} and \texpr{} and Single-Agent on \tsyn{}.

\subsection{Extended Scaling (Budget 32)}
\label{app:budget32}

To extend the scaling analysis beyond budget 16, we evaluate Multi-Agent with GPT-5-mini at 32 papers per gene in an extended scaling run (Table~\ref{tab:budget32}).

\begin{table}[ht]
\centering
\tablefont
\compacttablespacing
\caption{Multi-Agent (GPT-5-mini) at budget 16 vs.\ 32 in the extended scaling run.}
\label{tab:budget32}
\begin{tabular}{lccc}
\toprule
\textbf{Budget} & \textcolor{taskblue}{\textbf{GO R@30}} & \textcolor{taskgreen}{\textbf{Expr R@10}} & \textcolor{taskpurple}{\textbf{Syn R@20}} \\
\midrule
16 & 53.6\% & 58.5\% & 57.3\% \\
32 & 45.0\% & 50.7\% & 48.1\% \\
\bottomrule
\end{tabular}
\end{table}

In this run, \tgo{} recall decreases ($-8.6$pp), \texpr{} decreases ($-7.8$pp), and \tsyn{} decreases ($-9.2$pp) from budget 16 to 32. This supports the qualitative conclusion that simply expanding the paper budget is not enough: later papers can add noise, context-management pressure, and more opportunities for tool-use failures.

\subsection{Retrieval Method Ablation}
\label{app:retrieval}

To validate BM25~\citep{robertson2009probabilistic} as a baseline retrieval method, we compare four retrieval strategies under Single-Agent at budget 8 (Table~\ref{tab:retrieval-ablation}).

\begin{table}[ht]
\centering
\tablefont
\compacttablespacing
\caption{Retrieval method comparison (Single-Agent, budget 8, GPT-5-mini).}
\label{tab:retrieval-ablation}
\begin{tabular}{lccc}
\toprule
\textbf{Retrieval} & \textcolor{taskblue}{\textbf{GO R@30}} & \textcolor{taskgreen}{\textbf{Expr R@10}} & \textcolor{taskpurple}{\textbf{Syn R@20}} \\
\midrule
Hybrid (rerank) & \textbf{54.7\%} & 47.9\% & \textbf{58.1\%} \\
Hybrid (WRRF) & 52.6\% & \textbf{49.6\%} & 57.4\% \\
Lexical (BM25) & 50.5\% & \textbf{49.6\%} & 48.6\% \\
Semantic (MedCPT) & 23.9\% & 18.9\% & 27.8\% \\
\bottomrule
\end{tabular}
\tablenote{BM25 is lexical corpus retrieval; MedCPT is biomedical dense retrieval~\citep{jin2023medcpt}; Hybrid (rerank) denotes lexical-plus-semantic reranking; Hybrid (WRRF) uses weighted reciprocal-rank fusion~\citep{cormack2009reciprocal}.}
\end{table}

BM25 remains competitive on \tgo{} and \texpr{}, while hybrid retrieval offers additional gains, especially for \tsyn{}. Pure dense retrieval with MedCPT collapses, likely because scientific papers use precise gene symbols and terminology that lexical matching handles well. The released corpus supports any retrieval strategy, making retrieval an open research variable.

\subsection{Case Study: End-to-End Annotation of \textit{bsf}}
\label{app:case-study}

We trace a complete Single-Agent run for \textit{bsf} (bicoid stability factor, FBgn0284256) to illustrate the iterative search-read-refine loop and how \flybench{} evaluates the resulting annotations. The full run reads 16 papers over 6 search rounds in approximately 6 minutes (\$1.51 API cost).

\vspace{0.5em}
\noindent\textbf{Agent trace.}
Table~\ref{tab:case-study-trace} shows the full sequence of agent actions. The agent begins with broad queries, reads papers, discovers new vocabulary, and feeds it back into search---each round retrieving papers invisible to earlier queries.

\begin{table*}[ht]
\centering
\densetablefont
\densetablespacing
\caption{Single-Agent trace for \textit{bsf} (16 papers, 6 search rounds, \~{}6 min). Bolded terms in queries are vocabulary the agent learned from reading papers. The ``Notes'' column explains what each step achieves and what the agent discovers.}
\label{tab:case-study-trace}
\begin{tabular}{@{}p{0.09\textwidth}p{0.88\textwidth}@{}}
\toprule
\textbf{Action} & \textbf{Notes} \\
\midrule
\multicolumn{2}{l}{\textit{Phase 1: Initial retrieval}} \\
\midrule
\texttt{search} & Query: \texttt{"bicoid stability factor"} then \texttt{"bicoid stability factor bsf Drosophila"}. Broad keyword search using only the gene name. Returns papers that mention the gene but are not dedicated studies of it. \\
\midrule
\multicolumn{2}{l}{\textit{Phase 2: First reading pass --- 5 papers}} \\
\midrule
\texttt{read} & PMC3192837 --- The most informative paper. Describes the protein's subcellular location and role in mitochondrial RNA processing. The agent discovers that the gene has a \textbf{mammalian counterpart called LRPPRC} --- a name that never appears in the benchmark prompt. \\
\texttt{read} & PMC4663177 --- Shows the protein binds to RNA from another gene, \textbf{\textit{oskar}}. This is a new entity name the agent can search for. \\
\texttt{read} & PMC3529350 --- Detects the protein in muscle tissue. No new search terms learned. \\
\texttt{read} & PMC4503542 --- Describes a mutant strain called \textbf{\textit{SH1181}}. Papers studying this mutant may use the allele name instead of the gene name. \\
\texttt{read} & PMC4267620 --- Only mentions the gene in passing. Low information yield. \\
\midrule
\multicolumn{2}{l}{\textit{Phase 3: Refined search --- using \textbf{oskar} (learned from PMC4663177)}} \\
\midrule
\texttt{search} & Query: \texttt{Bicoid Stability Factor \textbf{oskar} BSF osk}. Adding the interaction partner retrieves papers about the oskar--bsf relationship that did not rank highly under gene-name-only queries. \\
\texttt{read} & PMC7001992, PMC8186867, PMC3174985 (3 papers). New papers found via refined query. \\
\midrule
\multicolumn{2}{l}{\textit{Phase 4: Refined search --- using \textbf{SH1181} (learned from PMC4503542)}} \\
\midrule
\texttt{search} & Query: \texttt{bsf \textbf{SH1181} "Bicoid Stability Factor"}. Adding the mutant allele name retrieves papers that study the gene through its mutant but may not mention the canonical gene name prominently. \\
\texttt{read} & PMC4918342, PMC10413924 (2 papers). \\
\midrule
\multicolumn{2}{l}{\textit{Phase 5: Refined search --- using \textbf{LRPPRC} (learned from PMC3192837)}} \\
\midrule
\texttt{search} & Query: \texttt{\textbf{LRPPRC} bsf Drosophila BSF \textbf{mitochondria}}. The highest-impact refinement. ``LRPPRC'' is the mammalian counterpart name that appears only in papers the agent already read --- not in the prompt or any external hint. This query retrieves \textbf{PMC6668790}$^\dagger$ and PMC5559520, which contain 4 ground-truth annotations between them. \\
\texttt{read} & PMC6668790$^\dagger$, PMC3474952, PMC10484445, PMC5624782, PMC8843987, PMC3250245 (6 papers). Fills the 16-paper budget. \\
\midrule
\multicolumn{2}{l}{\textit{Phase 6: Ontology resolution and submission}} \\
\midrule
\texttt{resolve} & 22 tool calls to map extracted text descriptions to standardized ontology identifiers (e.g., mapping ``located in mitochondria'' to the correct database term). \\
\texttt{submit} & Final output: 9 function annotations, 4 expression annotations, 5 synonyms. \\
\bottomrule
\multicolumn{2}{l}{\footnotesize $\dagger$ PMC6668790 contains 2 ground-truth annotations and was only reachable via the LRPPRC query.} \\
\end{tabular}
\end{table*}

\noindent Of the 16 papers read, \textbf{7 were found only by the later refined queries} (phases 3--5), and two contain ground-truth annotations missed by single-shot retrieval.

\vspace{0.5em}
\noindent\textbf{Evaluation.}
FlyBase curators have annotated \textit{bsf} with 8 unique in-corpus GO terms, 2 in-corpus expression annotations, and 6 in-corpus synonyms. The agent recovers 2 exact GO matches and 4 semantically close predictions (e.g., predicting ``sarcomere'' where curators annotated the more specific ``Z~disc''), 1 of 2 expression annotations, and all 6 synonyms. The synonym \textit{DmLRPPRC1}---accounting for 3 of 6 matches (case variants)---was itself discovered during reading, the same discovery that powered search step~16.

\vspace{0.5em}
\noindent\textbf{What the benchmark reveals.}
This case study illustrates two findings. First, the closed-loop search-read-refine behavior directly improves recall: vocabulary learned from reading (interaction partners, allele names, ortholog nomenclature) feeds back into search, retrieving papers that single-shot retrieval misses. This behavior is impossible to evaluate in benchmarks that provide pre-selected documents. Second, the dominant failure mode is \textit{ontological granularity mismatch}: the agent reads the right papers and identifies the correct biology, but maps evidence to ontology terms at the wrong specificity level. This recurs across genes and remains a systematic challenge for automated curation.

\end{document}